\documentclass[runningheads]{llncs}
\usepackage[T1]{fontenc}
\usepackage{graphicx}
\usepackage{graphicx}
\usepackage{booktabs}
\usepackage{multirow}
\usepackage{makecell}
\usepackage{algorithm}
\usepackage{algpseudocode}
\usepackage{amsmath,amsfonts,bm}
\usepackage{svg}
\usepackage{rotating}

\usepackage{mathtools}

\definecolor{MyGreen}{rgb}{0,0.6,0.3}
\definecolor{blackpink}{rgb}{0.6,0,0.6}
\definecolor{blood_red}{RGB}{200, 0, 0}

\begin{document}
%
% \title{Abrupt Change-Aware Time-series Forecasting via Local Discrepancy Density}
\title{Deep Imbalanced Time-series Forecasting via Local Discrepancy Density}
\titlerunning{Deep Imbalanced Time-series Forecasting via Local Discrepancy Density}
% \author{Anonymous authors}
\author{Junwoo Park \and
Jungsoo Lee \and
Youngin Cho \and  
Woncheol Shin \and Dongmin Kim \and Jaegul Choo \and Edward Choi}

\authorrunning{Park et al.}

\institute{Korea Advanced Institute of Science and Technology (KAIST), Daejeon, South Korea
\email{\{junwoo.park, bebeto, choyi0521, swc1905, tommy.dm.kim, jchoo, edwardchoi\}@kaist.ac.kr}}


\maketitle              % typeset the header of the contribution
\begin{abstract}
Time-series forecasting models often encounter abrupt changes in a given period of time which generally occur due to unexpected or unknown events. Despite their scarce occurrences in the training set, abrupt changes incur loss that significantly contributes to the total loss. Therefore, they act as noisy training samples and prevent the model from learning generalizable patterns, namely the normal states.
Based on our findings, we propose a reweighting framework that down-weights the losses incurred by abrupt changes and up-weights those by normal states. 
For the reweighting framework, we first define a measurement termed \emph{Local Discrepancy (LD)} which measures the degree of abruptness of a change in a given period of time. Since a training set is mostly composed of normal states, we then consider how frequently the temporal changes appear in the training set based on LD. 
%(\textit{i.e.}, estimated LD density).
% Since normal states generally appear frequently compared to the abrupt changes, they achieve higher LD density. Using such a property, 
% We reweight the losses proportionally to the estimated LD density. 
Our reweighting framework is applicable to existing time-series forecasting models regardless of the architectures. 
Through extensive experiments on 12 time-series forecasting models over eight datasets with various in-output sequence lengths, we demonstrate that applying our reweighting framework reduces MSE by 10.1\% on average and by up to 18.6\% in the state-of-the-art model.

\keywords{Time-series forecasting  \and Data imbalance \and Noisy samples.}
\end{abstract}
\section{Introduction}
\label{sec:intro}
As vast records are collected over time in diverse fields, the demand to predict the future based on the previous sequential data has led to efforts to solve the time-series forecasting problem in various applications such as energy~\cite{ahmad2014review}, economics~\cite{granger2014forecasting}, traffic~\cite{vlahogianni2014short}, weather~\cite{salman2015weather}, environment pollution~\cite{de2008field} and mechanical system monitoring~\cite{zhou2021informer}.
Previous studies focused on addressing the well-known challenges of time-series forecasting such as finding reliable dependencies from intricate and entangled temporal patterns~\cite{wu2021autoformer,oreshkin2019n} or extending the forecasting time (i.e., long-term forecasting)~\cite{zhou2022fedformer,liu2021pyraformer,wu2021autoformer,zhou2021informer}. 
For example, recent studies focused on improving the Transformer-based~\cite{vaswani2017attention} models to address the long-term forecasting by taking the advantage of the long-term capacity of the self-attention mechanism and reducing quadratic computational costs~\cite{li2019enhancing,zhou2021informer,wu2021autoformer,liu2021pyraformer}.

\begin{figure}[h!]
\begin{center}
  \includegraphics[width=1.0\linewidth]{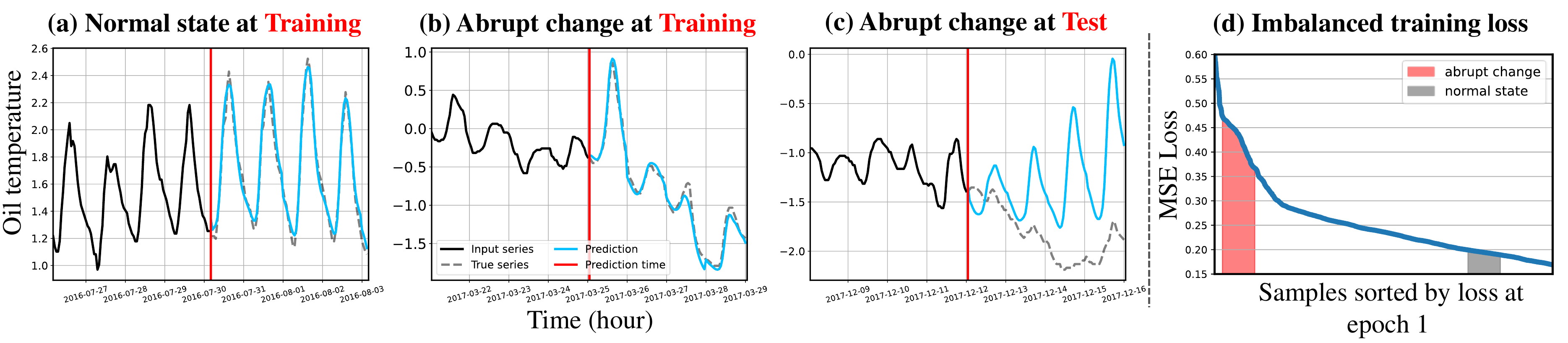}
\end{center}
\vspace{-5mm}
\caption{
% JW: (a) and (b) shows that the forecaster correctly predicts the target values during the training phase over both normal states (\textit{i.e.,} time stamps without abrupt changes) and abrupt changes, respectively.
We observe that the state-of-the-art forecaster correctly predicts the target values during the training phase over both (a) normal states and (b) abrupt changes, respectively.
However, (c) illustrates that the model fails to correctly predict the abrupt change during the test phase.
(d) shows imbalanced loss when the training samples are sorted by MSE loss of each sample in the early training phase. 
Our important finding is that the training samples with abrupt change (b) occupy the large portion of total loss. 
On the other hand, training samples within the normal states (a) have a relatively small loss.
This leads the model to focus less on the normal states during training.}
\label{fig:problem-fig}
\vspace{-5mm}
\end{figure}

Despite the remarkable improvements of the previous studies, even the state-of-the-art models take little account of the \emph{abrupt changes} in time-series data. 
Abrupt change refers to the drastic change of target values (either increase or decrease) beyond the extent of the changes observed in the recent past. 
These abrupt changes are challenging, if not impossible to predict based solely on previous observations of the target variable, as they are generally caused by unexpected and external events (\textit{e.g.}, natural disaster and war). 
Such changes break the auto-correlation structures, the periodic relationships between target variables, which are essential for a time-series forecaster to predict futures.
One straightforward remedy is to laboriously collect external variables (\textit{e.g.,} annotations of external events) and enforce a model to learn the relationship between the collected variables and the target variables (\textit{i.e.}, cross-correlation). 
However, utilizing additional variables without thorough verification causes the model to learn a spurious correlation between variables, which worsens the generalization ability.
Moreover, some abrupt changes have unknown causes (\textit{e.g.}, sensor malfunction), which cannot be addressed by simply collecting external variables.

While forecasting abrupt changes is known to be challenging~\cite{moniz2017resampling,hou2021deep}, even worse, another significant issue of abrupt changes is that they limit the generalization performance of forecasting models during the test phase.
Deep learning models are known to correctly predict all training samples regardless of the noisy labels by simply memorizing them (\textit{i.e.}, overfitting)~\cite{zhang2016understanding}.
Our finding is that recent time-series forecasting models can easily memorize even abrupt changes in which the output sequence shows the different temporal characteristics (\textit{e.g.}, mean, variance, and periodic structure) with the input sequence as shown in Figure~\ref{fig:problem-fig}.
To be more specific, Figure~\ref{fig:problem-fig}(a) and (b) show that the model correctly predicts the target values during the training phase in both normal states (\textit{i.e.}, trend or periodicity of input sequence maintained in the output sequence) and abrupt changes, respectively.
However, Figure~\ref{fig:problem-fig}(c) illustrates that the model fails to correctly predict the abrupt change during the test phase. 
The main reason is that the model is heavily overfitted to the abrupt changes since they take a significant portion of the total loss value compared to the ones in normal states (Figure~\ref{fig:problem-fig}(d)).
% \vspace{-5mm}
\begin{figure}[h!]
\includegraphics[width=1.0\linewidth]{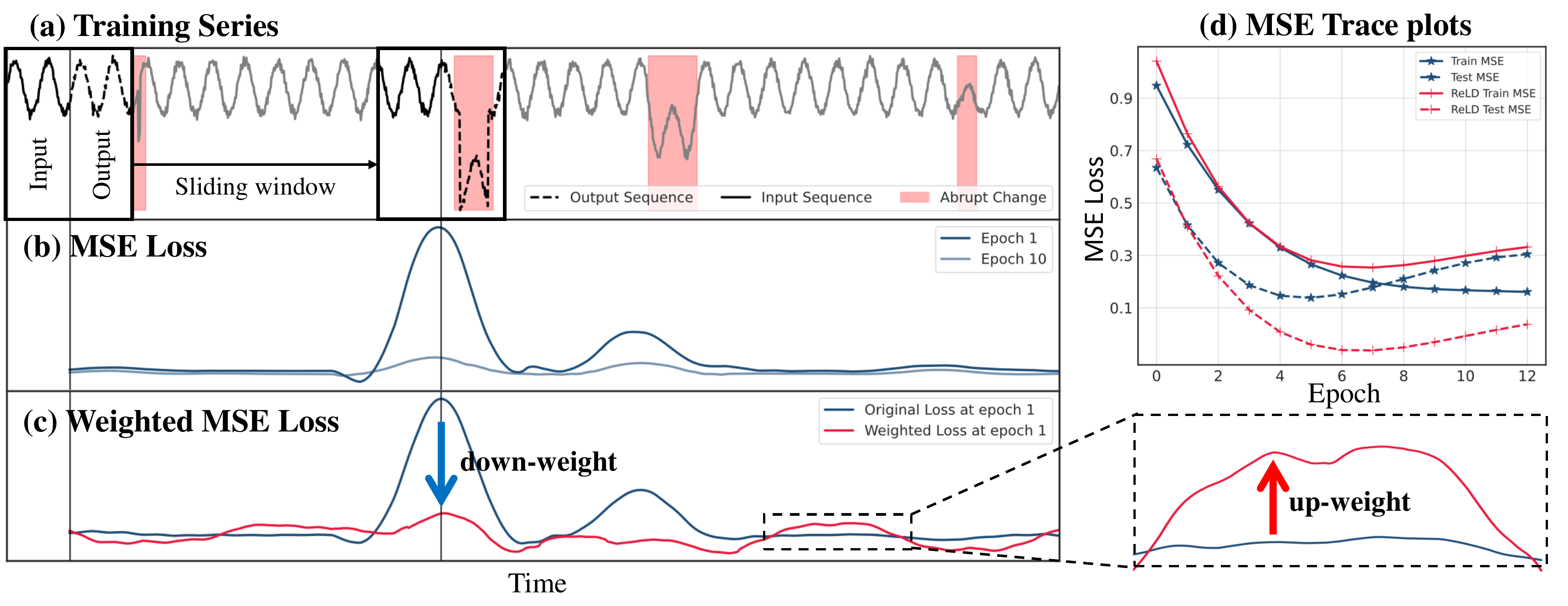}
\vspace{-7mm}
\caption{
(a) We trained a model with a training series including four abrupt changes (red-shaded regions).
(b) While the losses caused by the abrupt changes are considerably high in the early training phase, they are reduced significantly after several epochs of training.
(d) After the losses by abrupt changes are decreased, however, 
%Although the train loss decreased at a reasonable level, 
we observe that the test losses rather increase, implying the degraded generalization capability.
% (c) shows the re-weighted losses by our proposed framework that down-weights the loss of samples containing the abrupt changes and up-weight normal samples. 
(c) We mitigate such an issue by proposing a reweighting framework that down-weights the losses of samples containing the abrupt changes (blue arrow) and up-weights normal samples (red arrow).
% MSE loss plot (d) demonstrates that the model trained on our proposed re-weighting framework shows lower test MSE error compared to the model in existing framework.
(d) The model trained with our proposed reweighting framework achieves lower test MSE compared to that of the model without our framework.}
\label{fig:contri}
\vspace{-5mm}
\end{figure}
Therefore, we propose a simple yet effective reweighting framework that encourages the model to balance the imbalanced loss between abrupt changes and normal states.
Generally, time-series datasets do not provide explicit labels as to when the abrupt changes occur.
Moreover, explicitly bisecting time stamps into abrupt changes and normal states is challenging since the definition of abrupt change may be vague depending on perspectives. 
Thus, we define a measure called \emph{Local Discrepancy} (LD) which is used to determine how much a change in a given period of time is abrupt.
By sliding a fixed-size window over the training time-series data, we compute the statistical difference between the in-output sequences as LD.
Then, based on the observation that abrupt changes rarely appear in the training samples while normal states comprise the majority of the training set, we count the frequency of temporal changes based on LD. 
We divide the LD values into a predefined number of bins which are smoothed by kernel density estimation (\textit{i.e.}, estimated LD density).
By obtaining low LD density for the abrupt changes and high ones for the normal states, we \textbf{\emph{re}}weight loss values proportional to the estimated \textbf{\emph{L}}D \textbf{\emph{D}}ensity, which we term our method as \emph{ReLD}.
This enables to emphasize the normal states which are the ones a model should learn for enhancing the forecasting capability.
In summary, the main contributions of our work are as follows:
\begin{itemize}
    \item We reveal that the abrupt changes significantly degrade the time-series forecasting performance by taking most of the loss values.
    
    \item We propose a simple yet effective reweighting framework that adjusts the balance of the  loss based on LD density, namely ReLD.

    \item  Our reweighting framework consistently improves the performance of twelve existing time-series models on eight datasets, which reduces MSE by 10.1\% on average and up to 18.6\% when applied to the state-of-the-art model.

    \item ReLD also outperforms methods addressing the noisy samples such as smoothing, outlier filtering, and error-based baselines with a significant margin.
    
\end{itemize}

\section{Related Work}
\subsection{Deep Learning Models for Time-series Forecasting}
Deep learning-based models that have shown successful results in various domains have been actively applied to the time-series forecasting problem, which was originally dominated by classic statistical-based models~\cite{anderson1976time}. 
% Typically, previous time-series forecasting models leverage RNNs to learn the temporal dependencies inherent in time series~\cite{bahdanau2014neural}. 
% In addition, models using CNNs or those combined with RNNs have appeared to focus on the local characteristics of time series~\cite{bai2018empirical}.
Recent studies focused on extending forecasting time~\cite{zhou2022fedformer,liu2021pyraformer,wu2021autoformer,zhou2021informer}.
As the demand for long-term planning and early warning in the real-world applications has increased, long-term forecasting has become essential. 
Thus, transformer-based forecasting models, which are known to effectively learn global temporal patterns, have emerged.
These studies proposed sparse attention mechanisms to reduce the computational cost of the canonical attention mechanism when processing long sequences. 
% Another trend is to apply the time-series decomposition techniques to a deep learning model in an end-to-end training manner.
% While it was originally used for the time-series analysis, a model trained in such an approach is effective to capture the temporal dependencies which are considered as compositions of multiple signals. 
The previous studies have demonstrated their effectiveness on various time-series datasets across multiple domains.  
However, they do not deal with how the locally appearing anomalous patterns (\textit{i.e.}, abrupt changes) of time series affect the generalization capability of models.
\vspace{-3mm}
\subsection{Robustness Against Noisy Samples and Data Imbalance}
As aforementioned, deep learning models perfectly classify samples even with wrong annotations (\textit{i.e.}, noisy samples) by simply memorizing them during the training phase~\cite{zhang2016understanding}, an issue explored widely in image classification~\cite{co-teaching,gce}.
Similarly, the abrupt changes in time-series forecasting are generally occurred by unexpected or unknown events, making them challenging to forecast correctly solely based on the previous time series. 
Due to this fact, perfectly forecasting them during the training phase indicates that the models simply memorized them which are in fact noisy samples in the time-series data.
% Due to this fact, perfectly forecasting them during the training phase indicates that the models simply memorized them which are in fact noisy samples in the time-series data.
% Addressing such noisy samples has been explored widely in image classification~\cite{co-teaching,gce}, especially utilizing the fact that data samples with small loss are likely to be annotated with clean labels~\cite{noisy-memorization}. 
% For example, Generalized Cross-Entropy assigns less emphasis on the gradients of samples which show weak agreement between predicted labels and the ground truth labels since they are likely to be noisy samples~\cite{gce}.
% While reweighting approaches are effective to address noisy samples, such aforementioned studies are not originally designed to mitigate the data imbalance. 

% Addressing the data imbalance has been studied extensively in diverse fields~\cite{focal-loss,yang2021delving,decoupling}. 
% such as image classification~\cite{ldam,decoupling,HAR}, regression~\cite{ren2021bmse,yang2021delving}, object detection~\cite{focal-loss,imbalance-problem-object-detection,group-softmax}, and semantic segmentation~\cite{sml,cars-cant-fly}.
Unlike studies addressing noisy samples in other fields, the number of abrupt changes is excessively scarce compared to that of normal states in time-series, so considering the data imbalance in addition to the noisy samples is important.
The main intuition of addressing data imbalance is to emphasize the training of the minor samples based on the frequency of each class~\cite{focal-loss,yang2021delving}.
For example, Yang et al.~\cite{yang2021delving} proposed the label distribution smoothing method that addresses the data imbalance in the image regression task. 
To tackle such data imbalance in time-series forecasting due to the scarce temporal patterns, few studies proposed an augmentation approach~\cite{moniz2017resampling} or modified model architectures~\cite{hou2021deep}.
However, when addressing the data imbalance, they did not take account of models being overfitted to the scarce abrupt changes during the training phase.
In this regard, we propose a reweighting framework that takes both issues into account: 1) abrupt changes work as noisy samples, and 2) they cause the data imbalance. 
\vspace{-3mm}
\section{Method}
\label{sec:3-method}
\vspace{-2mm}
\subsection{Preliminary}
\label{sec:3-preliminary}
We first describe the forecasting task in a rolling window setting~\cite{li2019enhancing,zhou2021informer,wu2021autoformer,liu2021pyraformer}, which covers all possible in-output sequence pairs of the entire time series $\mathcal{S} = \{\mathbf{s}_1, \dots, \mathbf{s}_T \mid \mathbf{s}_t \in \mathbb{R}^m\}$, where $T$ is the length of observed series and $m$ denotes the number of variables at time $t$. 
Univariate and multivariate time-series forecasting addresses time-series data with $m=1$ and $m>1$, respectively. 
By sliding a fixed-size window on $\mathcal{S}$, we obtain the windows $\mathcal{D} = \{(\mathcal{X}_t, \mathcal{Y}_t)\}_{t=1}^{N}$, which are divided into two parts: input sequence $\mathcal{X}_t=\left\{\mathbf{s}_{t-I}, \ldots, \mathbf{s}_{t-1}\right\}$ with given length $I$ and output sequence $\mathcal{Y}_t=\left\{\mathbf{s}_{t}, \ldots, \mathbf{s}_{t+O-1} \right\}$ with length $O$ to predict. 
A forecaster $f$ predicts the most probable length-$O$ sequence in the future given the past length-$I$ sequence by learning temporal dependencies in $\mathcal{S}$. 
We mainly address the loss imbalance caused by the in-output sequence pairs which include a large discrepancy between adjacent $\mathcal{X}_a$ and $\mathcal{Y}_a$ compared to other $\mathcal{X}_t$ and $\mathcal{Y}_t$ pairs where $a$ is the time stamp with an abrupt change.
However, since most time-series datasets do not provide a label for the abrupt change, we propose a training framework in an unsupervised setting.
\begin{figure}[h!]
\begin{center}
  \includegraphics[width=1.0\linewidth]{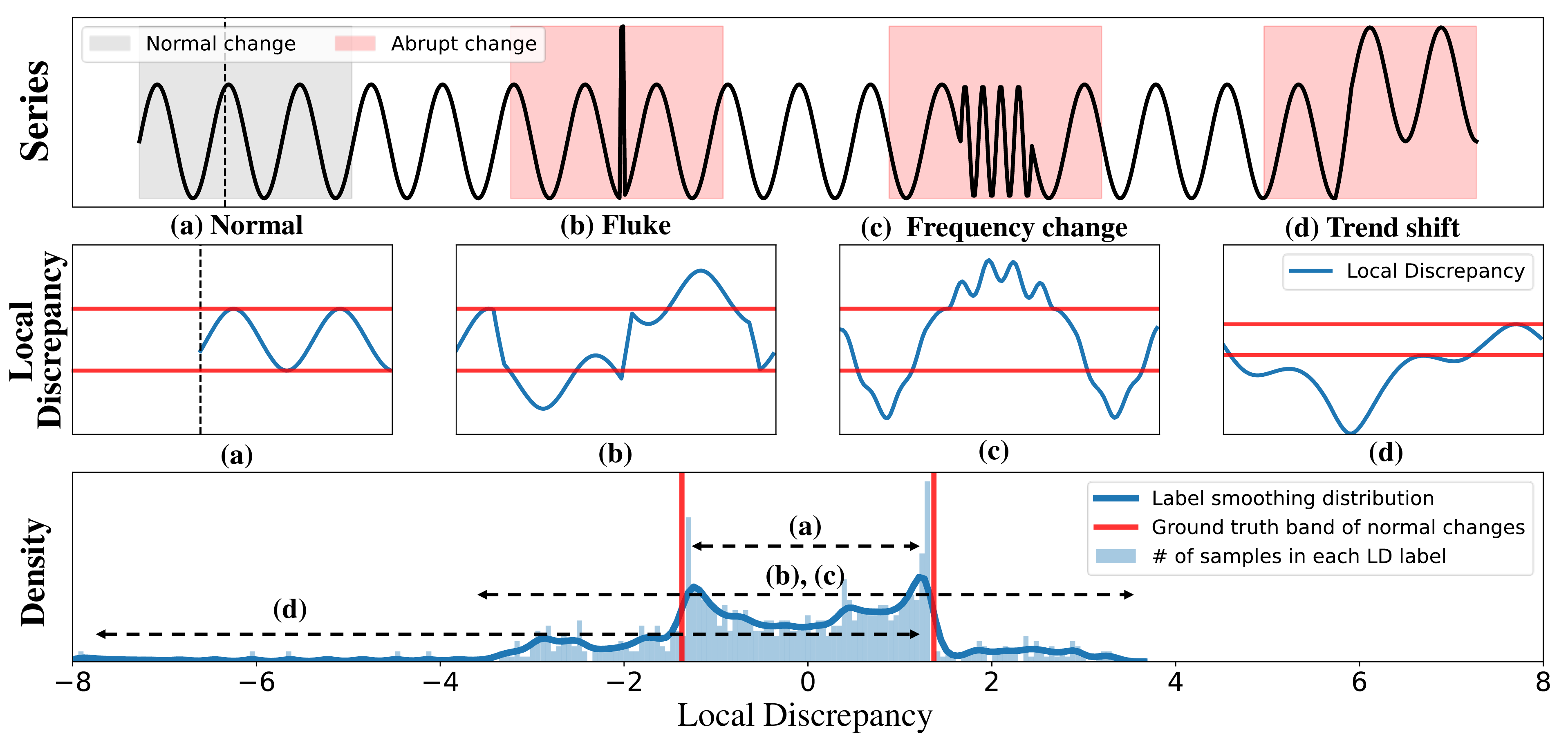}
\end{center}
 \vspace{-6mm}
   \caption{
   The four examples of temporal changes locally seen in time series data: (a) normal, (b) fluke, (c) frequency change, and (d) trend shift. 
   Local discrepancy computed by the sliding window captures the three abrupt changes beyond the bounds (red line) seen in normal states.
%   \js{While the red lines in second row indicate the maximum and minimum local discrepancies in normal states, the local discrepancies in abrupt changes ((b), (c), and (d)) exceed beyond the red lines.}
    In the estimated LD density distribution, training samples with abrupt changes are visibly fewer than training samples with normal state and are sparsely distributed with large absolute local discrepancy.
    % \js{In the estimated LD density distribution, we the training samples with abrupt changes are scarcely than training samples with normal state and are sparsely distributed with large absolute local discrepancy.}
   }
 \vspace{-3mm}
\label{fig:methodld}
\end{figure}
\vspace{-2mm}
\subsection{Local Discrepancy}
\label{sec:3-local-discrepancy}
We propose the \textit{Local Discrepancy} (LD) based on a statistical difference in order to measure how two adjacent in-output sequences, $\mathcal{X}_t$ and $\mathcal{Y}_t$, are different from each other.
We define LD as follows:
\begin{equation}
    \operatorname{LocalDis}(\mathcal{X}_t, \mathcal{Y}_t) = \frac{{\mathcal{\bar{X}}_t} - {\mathcal{\bar{Y}}_t}}{\sqrt{\frac{\mathrm{s}_{\mathcal{\bar{X}}_t}^2}{I} + \frac{\mathrm{s}_{\mathcal{\bar{Y}}_t}^2}{O} + \varepsilon}} \coloneqq v_t,
\end{equation}
where $\mathcal{\bar{X}}_t$ is the sample mean and $\mathrm{s}_{\mathcal{\bar{X}}_t}$ is the sample standard deviation of $\mathcal{X}_t$.

Statistical tests are generally used to determine whether means of two samples (\textit{i.e.}, groups of data points in a sequence) are identical or not~\cite{welch1938significance}. 
In this regard, we leverage t-statistic\footnote{Other statistics such as KPSS and $t$-squared can be used as LD. However, when we conduct a preliminary experiment, the $t$-statistic measures better than others. We further discuss the details in Section~\ref{sec:dislim}.}, a scalar value, as normalized discrepancy to measure how much two adjacent groups of samples are distinct. 
%%%% new version
Figure~\ref{fig:methodld} describes how LD reflects the different types of local temporal changes in time-series data (e.g., (a) normal changes, (b) fluke point, (c) frequency change, and (d) trend shift). The LD values of normal states oscillate within a certain range (see (a) red line) since the LD  also has periodicity as proven by Theorem~\ref{th1}, but the LD values of abrupt changes is beyond the range of normal LD. Additionally, the periodicity and boundedness of LD in normal periodic series are theoretically discussed in Supplementary~\ref{apx:threld} with all proofs.

\begin{theorem}[Periodicity of LD]
\label{th1}
\textit{If $f$ is a periodic function that satisfies $f(t) = f(t+p)$, 
\begin{equation}
    LD(a, a+L) = \frac{m(a) - m(a+L)}{\sqrt{\frac{s(a)}{N} + \frac{s(a+L)}{N}}}
\end{equation} is also a periodic function with period $p$, where $m(a) = \frac{1}{N}\sum_{t_\in I(a)}f(t)$, $s(a) = \frac{1}{N}\sum_{t_\in I(a)} (f(t) - m(a))^2 $, and $I(a) =\{a+\frac{L}{N}\cdot i\}_{i=0}^{N-1}$ for range $[a, a+L]$ and sampling interval $L/N$.}
\end{theorem}

As aforementioned, the definition of abrupt change may be vague depending on perspectives. Thus, rather than bisecting the time stamps into abrupt changes and normal states, we utilize LD values as weights of reweighting framework to mitigate the impact of abrupt changes in training phase. 
In other words, losses of training samples which have large absolute $v_t$ values will be down-weighed since we consider them to be close to the abrupt change.
By computing LD over the training dataset $\mathcal{D}_{train}$ and each of $m$ dimensions, we obtain the dataset $\mathcal{D}_{train} = \{(\mathcal{X}_t, \mathcal{Y}_t, v_t)\}_{t=1}^{N}$ containing local discrepancy $v_t \in \mathbb{R}^m$ for prediction time $t$ and for each of $m$ dimensions. We then assign the weight $w_t = \frac{c}{|v_t|+1} \propto \frac{1}{|v_t|+1} \in \mathbb{R}^m$ to each training sample inversely to LD value of sample in $\mathcal{D}_{train}$ with constant $c$ as scaling factor. Finally, we calculate the reweighted MSE loss $\mathcal{L}_w$ as follows:
% \vspace{-2mm}
\begin{equation}
\label{losseq}
\mathcal{L}_w(\mathcal{Y}_t, \mathcal{\hat{Y}}_t) = \frac{1}{m\cdot O} \cdot \sum_{j=1}^{m} w_t^j \sum_{i=0}^{O-1} \cdot (\mathbf{s}_{t + i}^{j} - \mathbf{\hat{s}}_{t+ i}^{j})^2
\end{equation} where $\mathcal{\hat{Y}}_t$ is forecasting results of $f$ conditioned on $\mathcal{X}_t$.
Through this simple reweighting framework which assigns weight inversely to LD values, namely \textit{invLD}, we down-weight the loss of abrupt changes (large absolute LD) and up-weight the loss of normal states (small absolute LD), following the observation that the original MSE loss in the presence of abrupt changes is much larger than the loss at the normal state. 
% This simple reweighting framework which assign weight to inversely LD values, namely iLD, improves the generalization ability of a model by mitigating the imbalanced loss values. 
Reweighting MSE only based on LD, however, does not take into account the property that normal states frequently appear while the abrupt changes are rarely included in the time-series data.
We further improve our reweighting framework by considering such frequency differences between abrupt changes and normal states. 

\subsection{Density-based reweighting for Time-series Forecasting}
\label{sec:3-density-reweight}
\vspace{-6mm}
\begin{figure}[h!]
\begin{center}
  \includegraphics[width=1.0\linewidth]{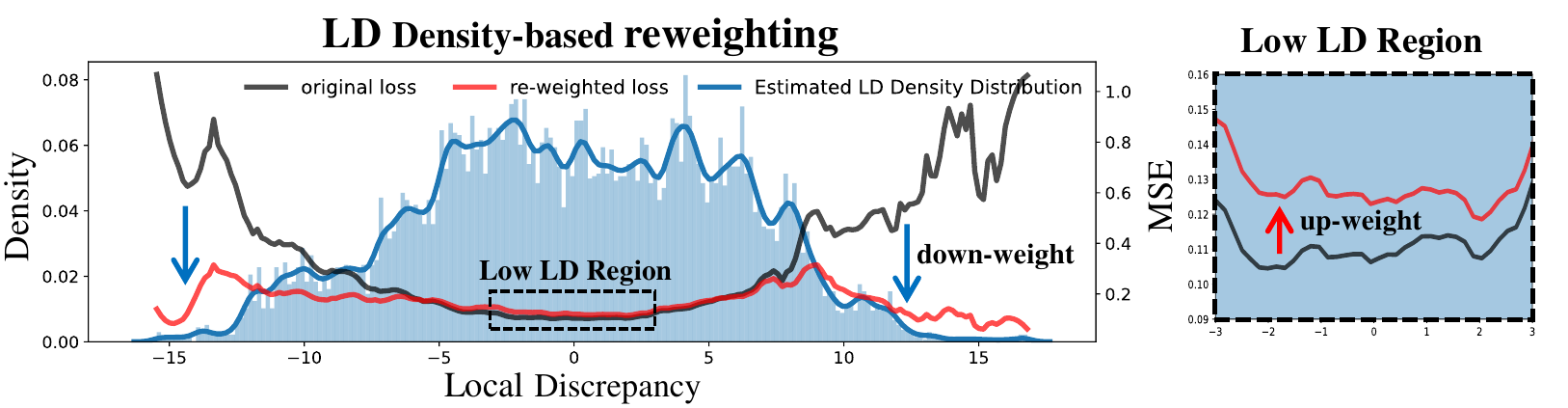}
\end{center}
\vspace{-5mm}
   \caption{For the real-world dataset (ETTh1), we visualize the estimated LD density distribution and the averaged MSE loss of samples in each LD bin after training a forecaster for one epoch.
   Our density-based re-weighting framework effectively down-weights (blue arrow) the losses on abrupt changes (low density and large LD) and up-weights (red arrow) those on normal states (high density and small LD).}
\vspace{-5mm}
\label{fig:density}
\end{figure}
Time series often exhibit both short-term and long-term repeating patterns~\cite{lai2018modeling} by periodicity, and taking them into account is crucial for making accurate predictions. 
Suppose a time series which has large shifts in a short period, but repeated. 
% Even if we assume such large shifts are part of the normal states, LD values within normal states have a large variance. 
We can assume such large shifts are part of the normal states considering their frequent occurrences. 
However, this temporal pattern is down-weighted because of their large LD values regardless of the number of occurrences.
In other words, invLD based on the inverse of LD (\textit{i.e.}, $w_t \propto \frac{1}{|v_t|}$ ) will not only down-weight the loss values of abrupt changes but also those of normal states, which the model should learn to properly forecast.
Therefore, we improve the time-series forecasting by considering the frequency of temporal changes (\textit{i.e.,} LD density) when reweighting loss values in time-series forecasting. 

Inspired by deep imbalanced regression~\cite{yang2021delving}, we use the kernel density estimation to address the missing regions between continuous LD spaces. 
Through the estimated density of LD $\tilde{p}(v)$, we assign the weight $w_t = c\cdot\tilde{p}(v_t) \propto \tilde{p}(v_t)$ and use these weights to train a model as $\mathcal{L}_w$ described in Equation~\ref{losseq}. 
Figure~\ref{fig:density} demonstrates that our final reweighting framework based on LD density, ReLD, mitigates the imbalanced loss problem in a real-world dataset. The procedure of our framework is described in Algorithm~\ref{algo:reld}.
\begin{algorithm}
\caption{\textbf{ReLD}: \textbf{Re}weighting framework based on \textbf{L}ocal Discrepancy \textbf{D}ensity}\label{alg:cap}
\begin{algorithmic}
\Require Training set $\mathcal{D}=\left\{\left(\mathcal{X}_{t}, \mathcal{Y}_{t}\right)\right\}_{t=1}^{N}$, bin size $\Delta b$, symmetric kernel distribution $k(v, v')$\\
Compute Local Discrepancy $\operatorname{LD}\left(\mathcal{X}_{t}, \mathcal{Y}_{t}\right)=\frac{\bar{\mathcal{X}}_{t}-\bar{\mathcal{Y}}_{t}}{\sqrt{\frac{s_{\bar{\mathcal{X}}_{t}}^{2}}{I}+\frac{s_{\bar{\mathcal{Y}}_{t}}^{2}}{O}+\varepsilon}} \coloneqq v_t$\\
Compute the empirical label density distribution $p(v)$ based on $\Delta b$ and $\mathcal{D}$ \\
Compute the effective label density distribution $\tilde{p}\left(v^{\prime}\right) \coloneqq \int_{\mathcal{V}} k\left(v, v^{\prime}\right) p(v) d v$

\ForAll{$\left(\mathcal{X}_{t}, \mathcal{Y}_{t},  v_t\right) \in \mathcal{D}$}
\State Assign weight for each sample as $w_{t} \propto c \cdot \tilde{p}\left(v_{t}\right)$ (constant $c$ as scaling factor)
\EndFor

\ForAll{number of training iterations}
\State Sample a mini-batch $\left\{\left(\mathcal{X}_{b},  \mathcal{Y}_{b}, w_{b}\right)\right\}_{b=1}^{B}$ from $\mathcal{D}$
\State Forward $\left\{\mathcal{X}_{b}\right\}_{b=1}^{B}$ and get corresponding predictions $\{\hat{\mathcal{Y}}_{b}\}_{b=1}^{B}$
\State Do one training step using the weighted loss $\frac{1}{B} \sum_{b=1}^{B} \mathcal{L}_{w_b}(\hat{\mathcal{Y}}_{b}, \mathcal{Y}_{b})$
\EndFor

\end{algorithmic}
\label{algo:reld}
\end{algorithm}

% ReLD: Re-weighting based on Local discrepancy Density 

% We illustrate the pseudo code of the ReLD in Algorithm~\ref{algo:reld}.

\vspace{-3mm}
\section{Experiments}
\label{sec:4-experiments}
\vspace{-3mm}
This section demonstrates that our proposed framework consistently improves existing time-series forecasting models regardless of the architectures.
% We apply our framework to 12 forecasting models and evaluate them on eight datasets which include four real-world forecasting applications: energy consumption, mechanical systems, weather, and pollution. 
Dataset analysis shows that our proposed framework brings larger performance gains as the number of abrupt changes in a given dataset increases. 
We also provide other experiments in the Supplementary, which include results on synthetic series, computational cost of methods, qualitative results, and details for reproducibility.
\vspace{-3mm}
\subsection{Experiment Setting}
\label{sec:4-experiment-setting}
\noindent \textbf{Dataset descriptions \enskip}
As mainstream benchmarks,  ETT are widely used to evaluate long-term forecasting methods~\cite{zhou2021informer,wu2021autoformer,liu2021pyraformer,zhou2022fedformer} 
ETT contains the crucial indicators (e.g., oil temperature, load, etc) collected from the electricity transformers over two years, and are categorized into four datasets depending on the location (ETT1 and ETT2) and interval (15 minutes and one hour). 
Electricity dataset contains the hourly electricity consumption of 321 customers from 2012 to 2014. 
Weather dataset is recorded every 10 minutes for a year, which contains 21 meteorological indicators (\textit{e.g.}, air temperature, humidity, etc). 
Pump dataset is collected from 52 sensors monitoring the water pump. 
AirQuality dataset~\cite{de2008field}, taken from the UCI repository, contains hourly averaged responses obtained from five metal oxide chemical sensors of an chemical multi-sensor. All dataset sources can be found in Supplementary.

\noindent \textbf{Forecasting models \enskip} 
% To verify that our reweighting framework works consistently in existing forecasting models, 
We applied it to 12 forecasting models and reported the reduced forecasting errors by applying ReLD.
% (i.e., counterpart)
The baselines are roughly categorized into three groups: Transformer-based~\cite{zhou2022fedformer,liu2021pyraformer,wu2021autoformer,zhou2021informer,kitaev2019reformer,li2019enhancing,vaswani2017attention}, CNN-based~\cite{bai2018empirical}, and RNN-based~\cite{bahdanau2014neural,lai2018modeling} models. 
We also include two univariate forecasting models: DeepAR~\cite{salinas2020deepar} and N-BEATS~\cite{oreshkin2019n}. 
% We applied our ReLD to the baselines without additional external data indicating abrupt changes or modifications of architectures.

% \noindent \textbf{Experiment details \enskip} 
% The forecasting performance is measured using the mean square error (MSE).
% Note that we follow the evaluation protocol proposed by previous work~\cite{zhou2021informer,wu2021autoformer}, including normalization through the mean and standard deviation of the train dataset.
% We conducted experiments on five different lengths of time series.
% All reported values are averaged MSE from five independent experiments with different random initializations. 
% % The results, which include confidence interval, are available in Appendix~\ref{}. 
% % %%%%%%%%%% Original submission %%%%%%%%%%%
% % Due to space limit, we only report the averaged MSE in the main paper while the confidence interval along with full benchmarks are included in Appendix~\ref{app:fullbench}. 
% % %%%%%%%%%%%%%%%%%%%%%%%%%%%%%%%%%%%%%%%%%%
% Due to the space limit, the main paper only reports the averaged MSE while the Supplementary includes the full benchmarks along with the confidence intervals.
\begin{table}[]
\caption{ Multivariate results with different input length $I$ and prediction lengths $O $. A lower MSE indicates a better prediction and the best results in each row are bolded. Imp. means averaged MSE reduction rate for a given model and dataset. Total denotes the averaged MSE reduction rate of a given dataset across all baselines models. The full results, which include other ETT datasets and confidence interval, are available in Supplementary~\ref{apx:fullbench}.}
\resizebox{\textwidth}{!}{
\begin{tabular}{@{}cccc|cc|cc|cc|cc|cc|cc|cc||c@{}}
\toprule
\multicolumn{2}{c}{Models}                                             & \multicolumn{2}{c}{FEDformer} & \multicolumn{2}{c}{Pyraformer} & \multicolumn{2}{c}{Autoformer}  & \multicolumn{2}{c}{Informer} & \multicolumn{2}{c}{Reformer}  & \multicolumn{2}{c}{LSTNet} & \multicolumn{2}{c}{LSTMa} & \multicolumn{2}{c||}{TCN} & \multirow{2}{*}{\rotatebox[origin=c]{90}{\textbf{Total}}} \\  

\cmidrule(lr){3-4}
\cmidrule(lr){5-6}
\cmidrule(lr){7-8}
\cmidrule(lr){9-10}
\cmidrule(lr){11-12}
\cmidrule(lr){13-14}
\cmidrule(lr){15-16}
\cmidrule(lr){17-18}

\multicolumn{2}{c}{I / O}                                             & base           & ReLD           & base         & ReLD         & base           & ReLD          & base          & ReLD          & base          & ReLD          & base            & ReLD           & base         & ReLD         & base         & ReLD       \\ \midrule

%ETTm1
\multicolumn{1}{l|}{\multirow{5}{*}{\rotatebox[origin=c]{90}{ETTm1}}} &  \multicolumn{1}{l|}{96/96} & 0.359 & \textbf{0.357} & 0.536 & 0.471 & 0.524 & 0.455 & 0.640 & 0.543 & 0.777 & 0.641 & 0.548 & 0.536 & 0.705 & 0.592 & 0.676 & 0.594  & \multirow{5}{*}{\rotatebox[origin=c]{90}{\textbf{-12.91\%}}}
 \\

\multicolumn{1}{l|}{}                       & \multicolumn{1}{l|}{336/168} &0.385 & \textbf{0.379} & 0.563 & 0.506 & 0.534 & 0.500 & 1.224 & 0.751 & 0.840 & 0.689 & 0.632 & 0.577 & 0.871 & 0.648 & 0.938 & 0.913   \\

\multicolumn{1}{l|}{}                       & \multicolumn{1}{l|}{336/336} & 0.403 & \textbf{0.396} & 0.697 & 0.573 & 0.561 & 0.514 & 1.390 & 1.008 & 0.987 & 0.895 & 0.798 & 0.686 & 1.125 & 0.681 & 1.148 & 1.126  
  \\

\multicolumn{1}{l|}{}                       & \multicolumn{1}{l|}{336/720} &0.501 & \textbf{0.480} & 0.904 & 0.682 & 0.560 & 0.528 & 1.333 & 1.078 & 1.122 & 1.003 & 0.925 & 0.817 & 0.978 & 0.828 & 1.277 & 1.238   \\  \cmidrule{2-18}

\multicolumn{1}{l|}{}                       & \multicolumn{1}{l|}{Imp.} & \multicolumn{2}{c|}{-1.99\%} & \multicolumn{2}{c|}{-16.17\%} & \multicolumn{2}{c|}{-8.42\%} & \multicolumn{2}{c|}{-25.11\%} & \multicolumn{2}{c|}{-13.88\%} & \multicolumn{2}{c|}{-9.16\%} & \multicolumn{2}{c|}{-24.14\%} & \multicolumn{2}{c|}{-4.95\%} \\  \midrule

%ETTm2
\multicolumn{1}{l|}{\multirow{5}{*}{\rotatebox[origin=c]{90}{ETTm2}}} &  \multicolumn{1}{l|}{96/96} & 0.189 & \textbf{0.184} & 0.371 & 0.248 & 0.293 & 0.221 & 0.445 & 0.286 & 0.743 & 0.449 & 0.443 & 0.343 & 0.381 & 0.280 & 0.554 & 0.384   & \multirow{5}{*}{\rotatebox[origin=c]{90}{\textbf{-21.42\%}}}
 \\

\multicolumn{1}{l|}{}                       & \multicolumn{1}{l|}{336/168} &0.343 & \textbf{0.275} & 0.566 & 0.551 & 0.309 & 0.283 & 2.283 & 1.453 & 1.208 & 0.836 & 0.950 & 0.830 & 1.178 & 0.601 & 1.868 & 1.956    \\

\multicolumn{1}{l|}{}                       & \multicolumn{1}{l|}{336/336} & 0.338 & \textbf{0.315} & 1.601 & 1.330 & 0.508 & 0.331 & 2.479 & 1.764 & 2.239 & 1.425 & 1.610 & 1.019 & 1.479 & 0.745 & 2.769 & 2.773 
  \\

\multicolumn{1}{l|}{}                       & \multicolumn{1}{l|}{336/720} & 0.432 & \textbf{0.393} & 5.476 & 5.037 & 0.502 & 0.413 & 6.580 & 5.777 & 3.068 & 2.827 & 6.130 & 4.449 & 3.083 & 2.381 & 3.204 & 3.187   \\  \cmidrule{2-18}

\multicolumn{1}{l|}{}                       & \multicolumn{1}{l|}{Imp.} & \multicolumn{2}{c|}{-9.47\%} & \multicolumn{2}{c|}{-15.15\%} & \multicolumn{2}{c|}{-21.39\%} & \multicolumn{2}{c|}{-28.29\%} & \multicolumn{2}{c|}{-28.65\%} & \multicolumn{2}{c|}{-24.85\%} & \multicolumn{2}{c|}{-36.96\%} & \multicolumn{2}{c|}{-6.57} \\  \midrule

%Weather
\multicolumn{1}{l|}{\multirow{5}{*}{\rotatebox[origin=c]{90}{Weather-h}}} & \multicolumn{1}{l|}{48/48}  & 0.338 & 0.336 & 0.292 & \textbf{0.279} & 0.344 & 0.343 & 0.345 & 0.294 & 0.343 & 0.313 & 0.318 & 0.310 & 0.346 & 0.325 & 0.348 & 0.327 
 &\multirow{5}{*}{\rotatebox[origin=c]{90}{\textbf{-4.41\%}}}\\

\multicolumn{1}{l|}{}                       & \multicolumn{1}{l|}{48/96}  &0.403 & 0.400 & 0.393 & \textbf{0.358} & 0.464 & 0.446 & 0.453 & 0.443 & 0.526 & 0.416 & 0.414 & 0.386 & 0.409 & 0.387 & 0.450 & 0.424      \\

\multicolumn{1}{l|}{}                       & \multicolumn{1}{l|}{96/192} &0.458 & 0.447 & 0.421 & \textbf{0.398} & 0.516 & 0.491 & 0.530 & 0.498 & 0.659 & 0.673 & 0.464 & 0.461 & 0.420 & 0.416 & 1.018 & 1.005  
  \\

\multicolumn{1}{l|}{}                       & \multicolumn{1}{l|}{168/336} &0.510 & 0.516 & 0.454 & \textbf{0.440} & 0.612 & 0.566 & 0.592 & 0.568 & 0.841 & 0.782 & 0.490 & 0.473 & 0.473 & 0.452 & 1.147 & 1.209     \\ \cmidrule{2-18}

\multicolumn{1}{l|}{}                       & \multicolumn{1}{l|}{Imp.}& \multicolumn{2}{c|}{-0.65\%} & \multicolumn{2}{c|}{-5.55\%} & \multicolumn{2}{c|}{-4.21\%} & \multicolumn{2}{c|}{-6.79\%} & \multicolumn{2}{c|}{-8.66\%} & \multicolumn{2}{c|}{-3.41\%} & \multicolumn{2}{c|}{-4.17\%} & \multicolumn{2}{c|}{-1.88\%}  \\  \midrule

%AirQuality
\multicolumn{1}{l|}{\multirow{5}{*}{\rotatebox[origin=c]{90}{AirQuality}}} &  \multicolumn{1}{l|}{96/96}  & 0.825 & \textbf{0.817} & 1.121 & 1.112 & 0.992 & 0.986 & 1.353 & 1.193 & 1.210 & 1.196 & 1.146 & 1.141 & 1.145 & 1.081 & 1.026 & 0.992   &\multirow{5}{*}{\rotatebox[origin=c]{90}{\textbf{-3.97\%}}}  \\

\multicolumn{1}{l|}{}                       & \multicolumn{1}{l|}{336/168} & 0.811 & \textbf{0.808} & 1.193 & 1.115 & 0.911 & 0.922 & 1.796 & 1.595 & 1.473 & 1.345 & 1.231 & 1.156 & 1.644 & 1.376 & 1.246 & 1.163    \\

\multicolumn{1}{l|}{}                       & \multicolumn{1}{l|}{336/336}  & 0.892 & \textbf{0.872} & 1.224 & 1.214 & 0.962 & 0.933 & 1.758 & 1.706 & 1.473 & 1.396 & 1.399 & 1.388 & 1.352 & 1.206 & 1.301 & 1.284  \\

\multicolumn{1}{l|}{}                       & \multicolumn{1}{l|}{336/720} & 0.997 & \textbf{0.953} & 2.196 & 1.982 & 1.129 & 1.079 & 2.914 & 2.985 & 1.723 & 1.671 & 1.826 & 1.921 & 2.475 & 2.333 & 1.442 & 1.426     \\  \cmidrule{2-18}

\multicolumn{1}{l|}{}                       & \multicolumn{1}{l|}{Imp.} & \multicolumn{2}{c|}{-2.01\%} & \multicolumn{2}{c|}{-4.47\%} & \multicolumn{2}{c|}{-1.72\%} & \multicolumn{2}{c|}{-5.87\%} & \multicolumn{2}{c|}{-4.51\%} & \multicolumn{2}{c|}{-0.52\%} & \multicolumn{2}{c|}{-9.63\%} & \multicolumn{2}{c|}{-3.09\%}  \\  \midrule

%Pump
\multicolumn{1}{l|}{\multirow{5}{*}{\rotatebox[origin=c]{90}{Pump}}} & \multicolumn{1}{l|}{96/96}  & 0.520 & \textbf{0.513} & 0.848 & 0.796 & 0.558 & 0.538 & 0.831 & 0.870 & 0.826 & 0.760 & 1.016 & 1.007 & 0.813 & 0.766 & 1.037 & 0.970   & \multirow{5}{*}{\rotatebox[origin=c]{90}{\textbf{-7.74\%}}}\\

\multicolumn{1}{l|}{}                       & \multicolumn{1}{l|}{336/168} & 0.550 & \textbf{0.536} & 0.851 & 0.843 & 0.597 & 0.581 & 1.705 & 1.527 & 1.094 & 0.856 & 1.327 & 1.202 & 0.909 & 0.816 & 1.109 & 1.077  
 \\

\multicolumn{1}{l|}{}                       & \multicolumn{1}{l|}{336/336} & 0.593 & \textbf{0.564} & 0.922 & 0.951 & 0.661 & 0.621 & 1.676 & 1.492 & 0.966 & 0.918 & 1.654 & 1.292 & 0.934 & 0.859 & 1.521 & 1.208   \\

\multicolumn{1}{l|}{}                       & \multicolumn{1}{l|}{336/720} & 0.723 & \textbf{0.580} & 1.370 & 1.283 & 0.707 & 0.619 & 1.704 & 1.699 & 1.328 & 1.218 & 1.608 & 1.642 & 1.464 & 1.244 & 2.075 & 1.546    \\  \cmidrule{2-18}

\multicolumn{1}{l|}{}                       & \multicolumn{1}{l|}{Imp.}& \multicolumn{2}{c|}{-7.10\%} & \multicolumn{2}{c|}{-2.61\%} & \multicolumn{2}{c|}{-6.17\%} & \multicolumn{2}{c|}{-4.28\%} & \multicolumn{2}{c|}{-10.73\%} & \multicolumn{2}{c|}{-7.49\%} & \multicolumn{2}{c|}{-9.77\%} & \multicolumn{2}{c|}{-13.85\%} \\

\bottomrule
\end{tabular}
}
\label{tab:main_multivariate}
\vspace{-5mm}
\end{table}

\vspace{-4mm}
\subsection{Main Results}
% \noindent \textbf{Multivariate results \enskip} 
As shown in Table~\ref{tab:main_multivariate}, applying our reweighting framework reduces the MSE consistently in all existing time-series forecasting models across different datasets and varying length-averaged settings. In addition, the lowest MSE in each setting was generally achieved by the models which applied ReLD. 
We also observe that the performance improvements vary depending on the datasets.  
For example, applying ReLD to the baselines achieves an average of 21.14\% lower MSE compared to the average of original errors on ETTm2.
On the other hand, applying ReLD achieves only 3.97\% lower MSE on average with AirQuality dataset.  
We analyze such an issue in Section~\ref{sec:dataset-analysis}. 
% \noindent \textbf{Univariate results \enskip} 
As for the univariate\footnote{Due to space limit, the univariate result are shown in Supplementary} setting,
% we list the univariate results of two typical datasets in Table~\ref{tab:main_univariate}. 
% Again, 
similar to the results observed in multivariate datasets, applying ReLD enhances the forecasting performance consistently regardless of the model architectures compared to baselines without ReLD. 
% Moreover, the lowest error for each setting was again achieved when ReLD was applied.
% \input{Tables/iclr_uni_main_results}

\vspace{-5mm}
\subsection{Comparisons with other methods}
\vspace{-2mm}
\noindent \textbf{Smoothing and outlier filtering methods \enskip}
Table~\ref{tab:comp-others} (a) compares our ReLD with two smoothing and outlier filtering methods.
Moving average (MA) and exponential MA (EMA) are widely used smoothing techniques that remove noisiness and reduce values of outliers, allowing meaningful temporal patterns to stand out. 
Similarly, outlier filtering also mitigates the influence of outliers on learning the normal patterns.
However, we observe that adopting such methods either shows insignificant performance improvement or rather aggravates the time-series forecasting performance. 
% This result shows that addressing the abrupt change is a challenging task and the performance improvement obtained by our proposed method is not trivial. 
% Details of the methods are described in Appendix~\ref{apx:smod}.

\noindent \textbf{Error-aware loss \enskip}
We also compare our method with error-based reweighting approaches for robust regression (L1, Huber~\cite{huber1992robust}, and IRLS~\cite{daubechies2010iteratively}) and data imbalance (Focal-R~\cite{focal-loss,yang2021delving} and flip Focal-R).  
Focal-R, the regression version of focal loss, allows a model to focus on samples with relatively large loss while down-weighting loss on samples with small errors.
It works in a way that is contrary to our findings.
We modified such an approach by putting negation on the input of Focal-R, termed as flip Focal-R (Details in Supplementary~\ref{apx:error-re}).
Table~\ref{tab:comp-others} (b) shows that the performance of Focal-R is rather degraded while that of flip Focal-R improved.
Such a result well demonstrates that our intuition, de-emphasizing the samples with high loss, is valid. 
Also, we observe that utilizing other error-based approaches fails to outperform our proposed method. 
We conjecture such superior performance of ReLD is mainly due to reflecting the temporal changes and periodicity.

\noindent \textbf{Ablation study of our reweighting framework \enskip} 
We conduct the ablation study of our proposed method by comparing our full framework ReLD and an approach which considers the LD values only (invLD).
Table~\ref{tab:comp-others} (c) shows that our full framework is superior to invLD.
Additionally, we observe that both approaches outperform the methods in (a) and (b).

\begin{table}[]
\caption{
Comparison with other methods which can deal with abrupt changes. We conduct experiments using ETTm2 dataset on two recent state-of-the-art time-forecasting models. 
% $\leftrightarrow$ means the replacement with original L2 loss and $+$ means the addition of method without the replacement.
`$\leftrightarrow$' indicates adopting the method in replace of the original L2 loss and `$+$' indicates adding the method to the original L2 loss.
}
\resizebox{\textwidth}{!}{
\begin{tabular}{c|lc|cccccc|cccccc|r}
\toprule
\multirow{4}{*}{\rotatebox[origin=r]{90}{Group}} & \multicolumn{2}{r|}{Models}                         & \multicolumn{6}{c}{FEDformer} & \multicolumn{6}{c|}{Autoformer} &  \\

\cmidrule(lr){4-9}
\cmidrule(lr){10-15}
% \cmidrule(lr){9-11}
% \cmidrule(lr){9}

&   &   \multicolumn{1}{r|}{I→O} &  \multicolumn{2}{c}{336 → 168}     &  \multicolumn{2}{c}{336 → 336}    &  \multicolumn{2}{c}{336 → 720}  & \multicolumn{2}{c}{336 → 168}     &  \multicolumn{2}{c}{336 → 336}    &  \multicolumn{2}{c|}{336 → 720}   &     \\ %\cmidrule(lr){2-10}
\cmidrule(lr){4-5}
\cmidrule(lr){6-7}
\cmidrule(lr){8-9}
\cmidrule(lr){10-11}
\cmidrule(lr){12-13}
\cmidrule(lr){14-15}
& Methods  &    & MSE & MAE &  MSE & MAE    &  MSE & MAE  & MSE & MAE  &  MSE & MAE    &  MSE & MAE   & Imp.      \\ \cmidrule(lr){2-16}
       
 & \multicolumn{2}{l|}{Vanilla (L2)}                        & 0.343 & 0.406 & 0.338 & 0.387 & 0.432 & 0.461 & 0.309 & 0.371 & 0.508 & 0.490 & 0.502 & 0.478  
  &  -   \\ \midrule
\multirow{3}{*}{(a)} &\multicolumn{2}{l|}{$+$ MA}                & 0.355 & 0.411 & 0.343 & 0.388 & 0.418 & 0.443 & 0.313 & 0.374 & 0.431 & 0.447 & 0.542 & 0.500 & -0.85\%   \\
&\multicolumn{2}{l|}{$+$ EMA}      &       0.364 & 0.419 & 0.343 & 0.389 & 0.404 & 0.432 & 0.319 & 0.377 & 0.516 & 0.473 & 0.549 & 0.506 & 1.33\% \\
&\multicolumn{2}{l|}{$+$ Outlier} &  0.292 & 0.364 & 0.330 & 0.380 & 0.405 & 0.429 & 0.384 & 0.422 & 0.420 & 0.444 & 0.468 & 0.475 & -3.26\%  \\
\midrule
 & \multicolumn{2}{l|}{$\leftrightarrow$ L1}                    & 0.282 & 0.345 & 0.321 & 0.366 & 0.402 & 0.416 & 0.308 & 0.368 & 0.349 & 0.391 & 0.434 & 0.439 & -11.24\% \\ 
\multirow{3}{*}{(b)} & \multicolumn{2}{l|}{$\leftrightarrow$ Huber}                    &0.285 & 0.353 & 0.322 & 0.369 & 0.418 & 0.432 & 0.307 & 0.369 & 0.398 & 0.424 & 0.452 & 0.456 & -8.32\% \\ 
 & \multicolumn{2}{l|}{$\leftrightarrow$ IRLS}                    & 0.281 & 0.345 & 0.322 & 0.368 & 0.398 & 0.416 & 0.292 & 0.356 & 0.350 & 0.387 & 0.435 & 0.433 & -12.12\% \\ 
 & \multicolumn{2}{l|}{$\leftrightarrow$ Focal-R}                    &0.403 & 0.451 & 0.377 & 0.423 & 0.504 & 0.523 & 0.315 & 0.379 & 0.445 & 0.463 & 0.520 & 0.497 & 6.02\% \\ 
&\multicolumn{2}{l|}{$\leftrightarrow$ flip Focal-R}         &0.284 & 0.344 & 0.322 & 0.367 & 0.405 & 0.417 & 0.307 & 0.366 & 0.371 & 0.405 & 0.470 & 0.453 & -9.70\% \\ 
\midrule

\multirow{2}{*}{(c)} &\multicolumn{2}{l|}{$+$ invLD } & 0.282 & 0.343 & 0.326 & 0.374 & 0.402 & 0.414 & 0.288 & 0.353 & 0.343 & 0.385 & \textbf{0.411} & \textbf{0.421} & -12.82\%  \\ 
&\multicolumn{2}{l|}{$+$ ReLD } & \textbf{0.275} & \textbf{0.339} & \textbf{0.315} & \textbf{0.361} & \textbf{0.393} & \textbf{0.409} & \textbf{0.283} & \textbf{0.348} & \textbf{0.331} & \textbf{0.377} & 0.413 & 0.422 & \textbf{-14.34}\%
 \\ 

\bottomrule
\end{tabular}
}
\label{tab:comp-others}
\vspace{-5mm}
\end{table}
\vspace{-6mm}
\subsection{Dataset Analysis}
\label{sec:dataset-analysis}

\noindent \textbf{Preserving the robustness on the abrupt changes \enskip} 
% Our reweighting framework improves time-series forecasting performance by mitigating the loss imbalance mainly occurred due to abrupt changes.
Since we impose less emphasis on the abrupt changes during the training phase, utilizing our framework may limit the model's ability to cope with the abrupt changes in the test phase.
% However, we confirm that our proposed framework does not degrade the forecasting performance on the abrupt changes.
Table~\ref{tab:abrupt} reports the MSE of test samples by categorizing them into time series with abrupt changes and those without abrupt changes.
For the experiment, we generated synthetic time-series dataset 
% from sinusoidal function of $t$ 
and injected abrupt changes into the series since the real-world dataset does not have labels for abrupt changes. 
% For this experiment, we use time series dataset generated from the sinusoidal function of $t$ which includes abrupt changes occasionally.
% We conducted the experiments with four recent state-of-the-art time-series forecasting models.
As originally intended, applying our framework achieves larger MSE reduction rates (\textit{i.e.}, MSE$_{N}$) compared to the ones without ReLD. 
As for the MSE of abrupt changes (\textit{i.e.}, MSE$_{A}$), the MSE$_{A}$ of three models decreased, and those of Pyraformer show competitive forecasting results. 
This result shows that our ReLD improves the forecasting performance on normal samples while preserving the robustness on the abrupt changes.
% This again demonstrates the general applicability of our work since it can be applied to a given dataset regardless of the existence of abrupt changes.

% \vspace{-3mm}
% \usepackage{multirow}
\begin{table}[]
% \caption{Forecasting results on synthetic time-series dataset in univariate setting.}
\caption{Forecasting results by categorizing time-series sequences into normal states and abrupt changes. 
%Averaged Imp. is the abbreviation of averaged improvements. 
We observe that our ReLD significantly reduces MSE on normal states (MSE$_N$) while also showing comparable MSE on abrupt changes (MSE$_A$).}
\resizebox{\textwidth}{!}{
\begin{tabular}{c|c|cc|cc|cc|cc|cc||rr}
\toprule
\multicolumn{2}{c|}{Prediction length}                  & \multicolumn{2}{c}{48} & \multicolumn{2}{c}{96} & \multicolumn{2}{c}{168} & \multicolumn{2}{c}{336} & \multicolumn{2}{c||}{720} & \multicolumn{2}{c}{Averaged Imp.} \\
% \cmidrule{1-12}
\cmidrule(lr){3-4}
\cmidrule(lr){5-6}
\cmidrule(lr){7-8}
\cmidrule(lr){9-10}
\cmidrule(lr){11-12}
\cmidrule(lr){13-14}

\multicolumn{1}{c|}{Model} & \multicolumn{1}{c|}{Metric} & Base       & ReLD      & Base       & ReLD      & Base       & ReLD       & Base       & ReLD       & Base       & ReLD       &    & Total                              \\ \midrule
\multirow{2}{*}{Pyraformer}                & MSE$_{N}$                 &    0.0702 & \textbf{0.0305} & 0.0580 & \textbf{0.0232} & 0.0547 & \textbf{0.0221} & 0.0449 & \textbf{0.0326} & 0.0379 & \textbf{0.0247}     &   \textbf{-47.67\%} & \multirow{2}{*}{-27.38\%}                                                  \\
                          & MSE$_{A}$                 &  0.4289 & \textbf{0.4168} & \textbf{0.5525} & 0.5905 & 0.6093 & \textbf{0.5882} & \textbf{0.4300} & 0.4350 & \textbf{0.2555} & 0.2665    &  1.21\%                                                 \\ \midrule
\multirow{2}{*}{Autoformer}                  & MSE$_{N}$                 &  0.2063 & \textbf{0.1473} & 0.2560 & \textbf{0.1430} & 0.2137 & \textbf{0.1261} & 0.3645 & \textbf{0.1857} & 0.5099 & \textbf{0.3950}    &    \textbf{-37.06\%} & \multirow{2}{*}{-33.25\%}                                                   \\
                          & MSE$_{A}$                &   0.7385 & \textbf{0.6946} & 0.9878 & \textbf{0.7312} & 0.8152 & \textbf{0.6541} & 0.8185 & \textbf{0.6224} & 0.8230 & \textbf{0.6776}  &     \textbf{-18.66\%}                                              \\\midrule
\multirow{2}{*}{N-BEATS}                        & MSE$_{N}$                &   0.0472 & \textbf{0.0345} & 0.0592 & \textbf{0.0401} & 0.0469 & \textbf{0.0355} & 0.0646 & \textbf{0.0411} & 0.0517 & \textbf{0.0394} 
   &         \textbf{-28.73\%}  & \multirow{2}{*}{-17.84\%}                                                \\
                          & MSE$_{A}$                 &    \textbf{0.3331} & 0.3794 & 0.6109 & \textbf{0.5375} & 0.5989 & \textbf{0.5944} & 0.4597 & \textbf{0.4197} & \textbf{0.2853} & 0.2909   &   \textbf{-1.12\%}                                          \\\midrule
\multirow{2}{*}{Informer}                        & MSE$_{N}$                 &   0.1350 & \textbf{0.0538} & 0.0819 & \textbf{0.0341} & 0.0762 & \textbf{0.0344} & 0.2954 & \textbf{0.0492} & 0.5564 & \textbf{0.1775}   &  \textbf{-64.96\%} & \multirow{2}{*}{-51.84\%}                                                \\
                          & MSE$_{A}$                &    0.5547 & \textbf{0.4746} & \textbf{0.5625} & 0.6031 & 0.6011 & \textbf{0.5810} & 0.7533 & \textbf{0.4694} & 0.7724 & \textbf{0.3619}   &  \textbf{-20.28\%}                      \\ 
\bottomrule
\end{tabular}
}
% \vspace{-6.5mm}
\label{tab:abrupt}
\end{table}
% \vspace{-5mm}
\begin{figure}[h!]
\begin{center}
  \includegraphics[width=1.0\linewidth]{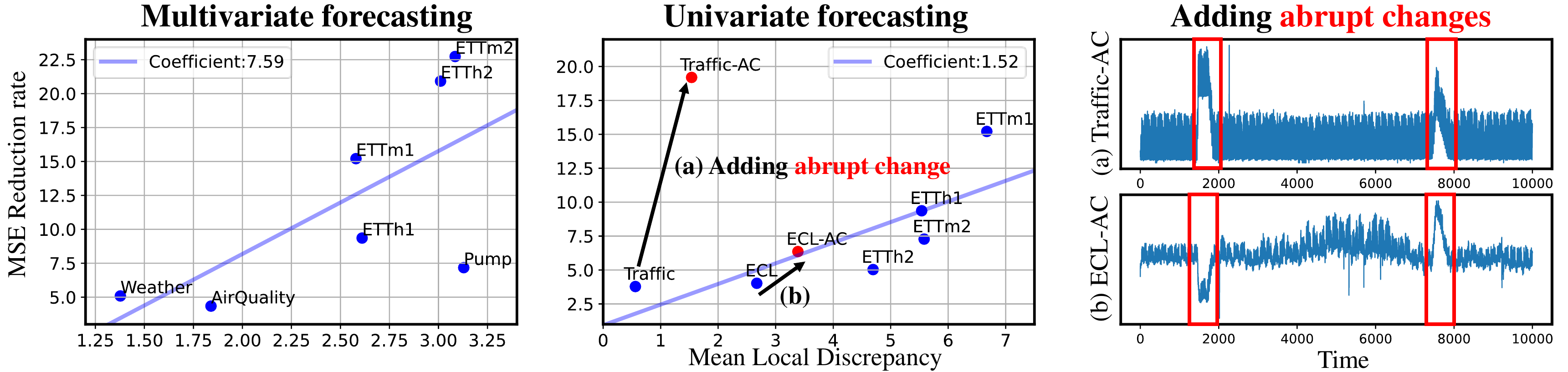}
\end{center}
\vspace{-5mm}
   \caption{
   Scatter plots showing the correlation between the averaged LD of each dataset and MSE reduction rates of experiments on the multivarite and univariate settings.
%   We observed that there exists a positive linear correlation between LD and the MSE reduction rate. 
%   To further demonstrate such a finding, we intentionally inject the abrupt changes highlighted as the red boxes in (a) and (b) into the traffic and ECL datasets, respectively.
  }
\label{fig:scatter-mse-ld}
\vspace{-5mm}
\end{figure}
\noindent \textbf{Different performance gains across datasets \enskip}
From the multivariate results (Table~\ref{tab:main_multivariate}) and univariate results,
%(Table~\ref{tab:main_univariate})
we found that the reduction rates of MSE vary depending on the datasets. 
For an in-depth analysis, we present the correlation between the average of reduction rate and the average of LD for each dataset using the scatter plot in Figure~\ref{fig:scatter-mse-ld}.
We observed that there exists a positive linear correlation between LD and the reduction rate, indicating that we obtain a higher reduction rate of MSE as the average of LD increases in a given dataset.
To further demonstrate such a finding, we intentionally inject abrupt changes into the Traffic and ECL, the datasets which showed the marginal improvements in the univariate setting.
We obtained a larger reduction rate of MSE with both Traffic and ECL including intentional abrupt changes compared to the original datasets.
This demonstrates that the marginal performance gain in both Traffic and ECL is due to the few number of abrupt changes in the dataset. Note that datasets without such abrupt changes might be well estimated with existing time-series forecasting models.
However, we emphasize that using ReLD does not degrade performance on such datasets, if not marginally improve it, due to a few number of abrupt changes inevitably included in time-series datasets.
% On the other hand, throughout the paper, we demonstrated that datasets including a non-trivial number of abrupt changes need improvements and using ReLD effectively improves the time-series forecasting performance on such datasets. 
% Abrupt changes in time-series data is unknown beforehand and our proposed framework does not require whether there exists such changes in the dataset. 
% In this perspective, applying ReLD as default on existing models improves the time-series forecasting performance without prior knowledge on the abrupt changes in a given dataset. 

\subsection{Computational cost of ReLD}
\label{apx:cost}
% Our reweighting framework requires the additional computational cost of calculating the weights for all input-output sequences before training. 
% However, as shown in Table~\ref{tab:cost}, the cost of calculating the weights on datasets with multiple settings is less than 1\% of the time it takes to train with the dataset during one epoch. 
Our reweighting framework requires a marginal amount of additional computational cost of calculating the weights for all input-output sequences before training. 
As shown in Table~\ref{tab:cost}, the cost of calculating the weights on datasets with multiple settings is less than 1\% of the time it takes to train with the dataset during one epoch.
The absolute time was mostly less than 1 second.

\begin{table}[]
\caption{The processing time of ReLD and training time of Autoformer during 1 epoch.}
\resizebox{\textwidth}{!}{
\begin{tabular}{@{}cccccc|c@{}}
\toprule
 &  \# of Windows & Window size  & \# of Series & ReLD Preprocessing time (a) & Training time (b)  & Ratio \\ 
Dataset &  & (I + O) &  & (seconds) & (seconds per epoch) &   (a) / (a) + (b) \\
\midrule
ETTh1	& 8449	& 192 (96 + 96) & 	7	& 0.18 & 38.12 &	0.47\%  \\
ETTh2	&8449	&192 (96 + 96)	&7	&0.17	&39.11	&0.43\%  \\
ETTm1	&34369	&192 (96 + 96)	&7	&0.69	&154.68	&0.44\%  \\
ETTm2	&34369	&192 (96 + 96)	&7	&0.68	&160.31	&0.42\%  \\
Weather-hour	&5093	&1056 (336 + 720)	&21	&0.63	&121.19	&0.52\%  \\
Pump	&9610	&672 (336 + 336)	&35	&1.22	&125.30	&0.96\%  \\
ECL	&17741	&672 (336 + 336)	&1	&0.08	&94.67	&0.08\%  \\
Traffic	&11225	&1056 (336 + 720)	&1	&0.07	&99.05	&0.07\%  \\

\bottomrule
\end{tabular}
}
\label{tab:cost}
\end{table}

\section{Discussion \& Limitation}
\vspace{-2mm}
\label{sec:dislim}
In this paper, we reveal that abrupt changes between adjacent sequences deteriorate the generalization performance of time-series forecasting models by occupying most of the losses despite their scarce occurrence in the training set. 
To solve this problem, we propose a simple yet effective reweighting framework that down-weights loss values of abrupt changes and up-weights those of normal states based on LD density.
Although our ReLD consistently enhances the performance on real-world datasets, there is a limitation we found.
We assume that an abrupt change is caused by unobserved external variables. 
However, if we can have access to those variables, our framework may not show performance improvement from the down-weighted losses of the abrupt changes. 
% We further discuss this limitation in Appendix~\ref{sec:externel}. 
% % Another limitation is failing to provide one best way to formulate LD, as it can be done with any statistics that can capture temporal changes. 
% Another limitation is that LD may be formulated with other statistics that can capture temporal changes.
% We report the results of balancing MSE using other statistics (e.g., $t$-squared for multivariate test and KPSS~\cite{kwiatkowski1992testing} for a stationarity test) in Appendix~\ref{apx:variousld} which shows that utilizing t-statistics (the statistic used in our work) outperforms them. Rather than using predefined statistics to identify abrupt changes, training learnable parameters for detecting them similarly done in time-series anomaly detection~\cite{malhotra2016lstm,audibert2020usad,xu2021anomaly} would be another way to balance MSE against abrupt changes.

\bibliographystyle{splncs04}
\bibliography{references}

\clearpage
\appendix
\section{Theoretical analysis of ReLD}
\label{apx:threld}
% As shown in Figure 3, we can observe the temporal patterns in the time-series and corresponding LD values oscillate within a certain range. 
As shown in Figure 3 of main paper, we can observe LD values oscillating within a certain range similarly to original time series. 
We discuss the following points about this observation. 
\begin{enumerate}
    \item When the time series is sampled from a periodic function, LD is also a periodic function.
    \item When the time series is sampled from a bounded periodic function, LD is bounded.
\end{enumerate}

First, we can easily prove that LD is also periodic when the time series is sampled from a periodic function, which the model should learn from data (i.e., normal states). 

 \paragraph{Theorem 1.} \textit{If $f$ is a periodic function that satisfies $f(t) = f(t+p)$, 
\begin{equation}
    LD(a, a+L) = \frac{m(a) - m(a+L)}{\sqrt{\frac{s(a)}{N} + \frac{s(a+L)}{N}}}
\end{equation} is also a periodic function with period $p$, where $m(a) = \frac{1}{N}\sum_{t_\in I(a)}f(t)$, $s(a) = \frac{1}{N}\sum_{t_\in I(a)} (f(t) - m(a))^2 $, and $I(a) =\{a+\frac{L}{N}\cdot i\}_{i=0}^{N-1}$ for range $[a, a+L]$ and sampling interval $L/N$.}

To prove Theorem 1, we prove and use Proposition 1 and 2 with respect to the mean and the variance of the periodic function. 

\paragraph{Proposition 1.} \textit{If $f$ is a periodic function that satisfies $f(t) = f(t+p)$, $m(a) = \frac{1}{N}\sum_{t_\in I(a)} f(t) $ is also a periodic function with period $p$ where $I(a) =\{a+\frac{L}{N}\cdot i\}_{i=0}^{N-1}$ for range $[a, a+L]$ and sampling interval $L/N$.}

\paragraph{Proposition 2.} \textit{If $f$ is a periodic function that satisfies $f(t) = f(t+p)$, $s(a) = \frac{1}{N}\sum_{x_\in I(a)} (f(t) - m(a))^2 $ is also a periodic function with period $p$ where $I(a) =\{a+\frac{L}{N}\cdot i\}_{i=0}^{N-1}$ for range $[a, a+L]$ and sampling interval $L/N$.}

By proposition 1 and 2, we prove Theorem 1 as follows: 
\begin{align*}
    LD(a+p, a+p+L) &= \frac{m(a+p) - m(a+p+L)}{\sqrt{\frac{s(a+p)}{N} + \frac{s(a+p+L)}{N}}} \\
                   &= \frac{m(a) - m(a+L)}{\sqrt{\frac{s(a)}{N} + \frac{s(a+L)}{N}}} \\
                   &= LD(a, a+L)
\end{align*}

% For the bound of LD, if we define LD as the difference between mean of input and mean of output (i.e., $LD(a, a+L) = m(a) - m(a+L)$), we can obtain the bound $2\cdot B$ given that $f$ is  bounded function that satisfies $|f(t)| \leq B$ for all $t$.
Regarding the bound of LD, if we define LD by ignoring the variances (i.e., $LD(a, a+L) = m(a) - m(a+L)$), we can obtain the bound $2\cdot B$ given that $f$ is  bounded function that satisfies $|f(t)| \leq B$ for all $t$.

However, since we use the variance of input and output, LD can diverge when both variances of input sequence and output sequence are equal to zero. There are two cases when the variance equals to zero.
\begin{enumerate}
    \item $f$ is a constant function.
    \item The window size $L$ is $N\cdot p$ for data points, which are sampled from a periodic function $f$ with the period $p$ and sampling interval $L/N$.
\end{enumerate}

In the first case, since the time-series dataset has a constant target value, the prediction also remains as the constant value, leading to a trivial solution. In the second case, the variance is no longer zero if the window size $L$ is adjusted.

In practice, we use epsilon $\epsilon$ as a numerical stabilizer to solve the case where variances are zero as shown in Equation 1. 
% Furthermore, we can assume that most real-world time-series contain white noise [1, 2, 3], which have time-independent variance. 
Note that we do not use these bounds as thresholds in the proposed method.

Proofs for the Proposition 1 and 2 are as follows.
\paragraph{Proposition 1.} \textit{If $f$ is periodic function that satisfy $f(t) = f(t+p)$, $m(a) = \frac{1}{N}\sum_{t_\in I(a)} f(t) $ is also periodic function with period $p$ where $I(a) =\{a+\frac{L}{N}\cdot i\}_{i=0}^{N-1}$ for range $[a, a+L]$ and sampling interval $L/N$.}

\paragraph{Proof:}
\begin{align*}
    m(a + p) &= \frac{1}{N}\sum_{t_\in I(a + p)} f(t) \\
             &= \frac{1}{N}\left\{f(a + p) + f(a + p + \frac{L}{N}) + \dots + f(a+ p + L - \frac{L}{N})\right\} \\
             &= \frac{1}{N}\left\{f(a) + f(a + \frac{L}{N}) + \dots + f(a + L - \frac{L}{N})\right\} \\
             &= m(a)
\end{align*}

\paragraph{Proposition 2.} \textit{If $f$ is periodic function that satisfy $f(t) = f(t+p)$, $s(a) = \frac{1}{N}\sum_{t_\in I(a)} (f(t) - m(a))^2 $ is also periodic function with period $p$ where $I(a) =\{a+\frac{L}{N}\cdot i\}_{i=0}^{N-1}$ for range $[a, a+L]$ and sampling interval $L/N$.}

\paragraph{Proof:}
\begin{align*}
    s(a + p) &= \frac{1}{N}\sum_{t_\in I(a + p)} \left\{f(t) - m(a + p)\right\}^2 \\
             &= \frac{1}{N}\left\{(f(a + p) - m(a + p))^2 + (f(a + p + \frac{L}{N}) - m(a + p))^2 \right. \\
             &\left. + \dots + (f(a+ p + L - \frac{L}{N}) - m(a + p))^2\right\}  \\
             &= \frac{1}{N}\left\{(f(a) - m(a))^2 + (f(a + \frac{L}{N}) - m(a))^2 + \dots + (f(a + L - \frac{L}{N}) - m(a))^2\right\}  \\
             &= s(a)
\end{align*}

% \section{Analysis ReLD}
\section{In-Depth Analysis on ReLD}
This section provides various analysis for our reweighting framework.   
\subsection{Abrupt change with external variables}
\label{sec:externel}
We assumed that an abrupt change can be caused by unobserved and external events as we mentioned in Section~\ref{sec:intro} and Section~\ref{sec:dislim}.
If the abrupt change can be predicted using an external variable, down-weighting the loss of the abrupt change would get in the way of learning such correlation for the model. However, utilizing additional variables without thorough verification causes the model to learn a spurious correlation between variables, which worsens the generalization ability. Moreover, some abrupt changes have unknown causes (e.g., sensor malfunction), which cannot be addressed by simply collecting external variables. In fact, as shown in the Table~\ref{tab:external}, we observed that training baseline models with external variables (i.e., Multivariate to Univariate denoted as Mul2Uni setting) rather shows lower performance than training those with the target time series only (i.e., Univariate to Univariate denoted as Uni2Uni setting). 
% These results at least indicate that simply adding covariates is not always a performance-enhancing solution. 
These results indicate that simply adding covariates does not guarantee performance gains. 
Note that multivariate forecasting we mentioned in main paper is multivariate to multivariate setting (i.e., Mul2Mul), which is different with Mul2Uni setting. In addition, applying our method on Mul2Uni outperformed the Uni2Uni in several cases (see Autoformer 96/96 and 336/168 of Table~\ref{tab:external}).

\begin{table}[]
\caption{
Comparison between Mul2Uni and Uni2Uni forecasting on ETTm1 dataset.
}
\resizebox{\textwidth}{!}{
\begin{tabular}{lc|ccc|ccc}
\toprule
 & Model                         & \multicolumn{3}{c}{Pyraformer} & \multicolumn{3}{c}{Autoformer}  \\

\cmidrule(lr){3-5}
\cmidrule(lr){6-8}
% \cmidrule(lr){9-11}
% \cmidrule(lr){9}

       Setting & I → O                          & 96 → 96     & 336 → 168     & 336 → 336  & 96 → 96     & 336 → 168     & 336 → 336          \\ \midrule
\multicolumn{2}{l|}{Uni2Uni}                        & 0.0821 ± \small0.0289 & 0.1286 ± \small0.0346 & 0.1941 ± \small0.0523 & 0.0577 ± \small0.0081 & 0.0881 ± \small0.0284 & 0.0903 ± \small0.0096      \\
\multicolumn{2}{l|}{Uni2Uni + Ours}                 &  \textbf{0.0576} ± \small0.0079 & \textbf{0.1218} ± \small0.0395 & \textbf{0.1843} ± \small0.0507 & 0.0522 ± \small0.0035 & 0.0723 ± \small0.0068 & \textbf{0.0847} ± \small0.0084  \\
\multicolumn{2}{l|}{Mul2Uni} &  0.1757 ± \small0.0372 & 0.2926 ± \small0.0778 & 0.5920 ± \small0.0591 & 0.0619 ± \small0.0090 & 0.0799 ± \small0.0188 & 0.1367 ± \small0.0392    \\
\multicolumn{2}{l|}{Mul2Uni + Ours}                    & 0.1137 ± \small0.0295 & 0.2984 ± \small0.1246 &0.5533 ± \small0.0761 & \textbf{0.0496} ± \small0.0021 & \textbf{0.0674} ± \small0.0071 & 0.1193 ± \small0.0267    \\ 
\bottomrule
\end{tabular}
}
\label{tab:external}
\end{table}

\subsection{ReLD on repeated changes}
\label{apx:rect}
\begin{figure}[h!]
\begin{center}
  \includegraphics[width=\linewidth]{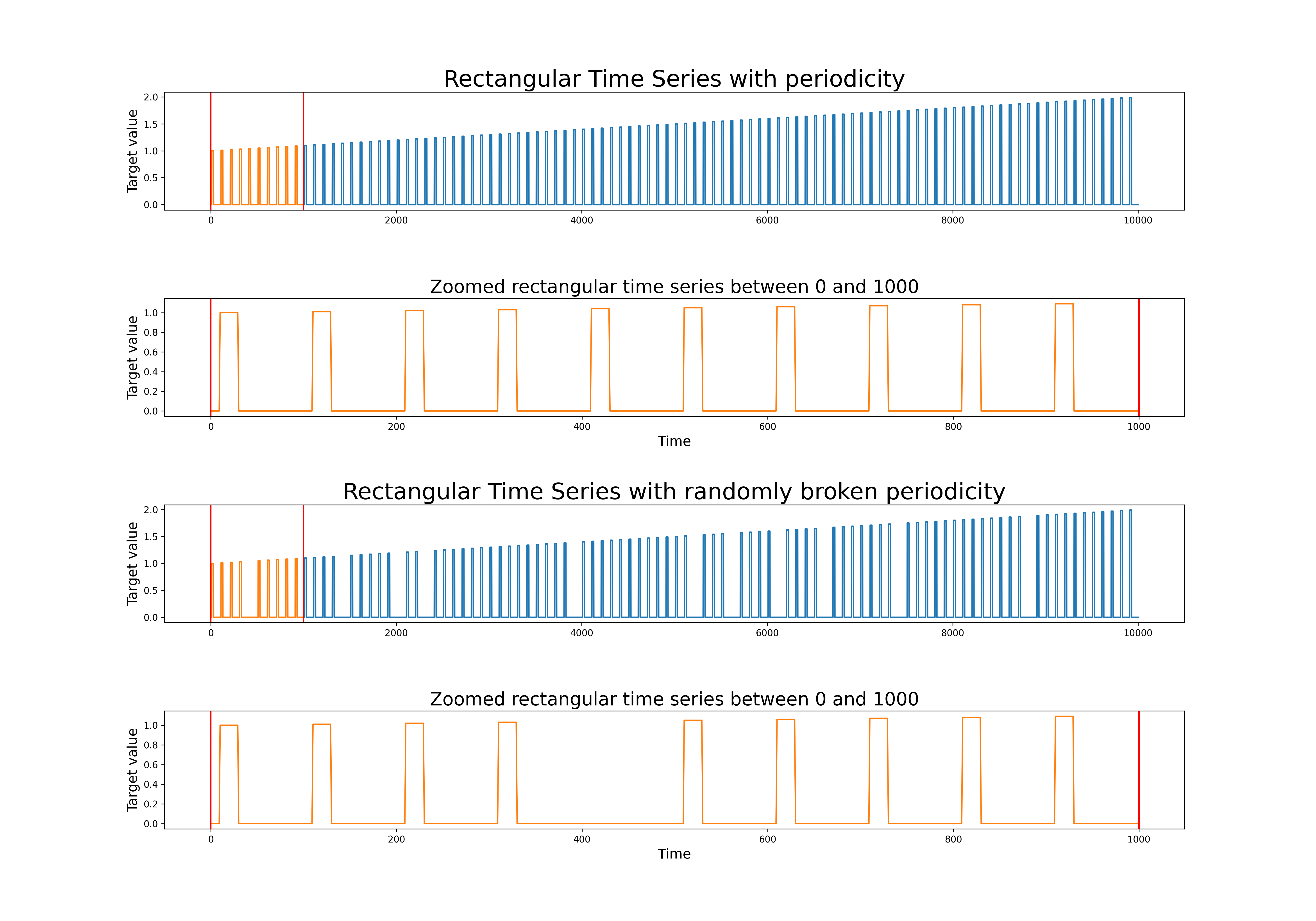}
\end{center}
\vspace{-5mm}
\caption{Two rectangular time series which include large shift in a short time. 
As in the first row, if a rectangular pattern with a large change exists several times, ReLD learns it normally without down-weighting it. However, if the periodicity is broken by sensor malfunctions in the third row, ReLD mitigates impact of anomaly pattern in training phase.}
\label{fig:suple-rect}
\end{figure}
To further understand our ReLD, we present a rectangular time series as a special case, which generally includes a large shift during a short period of time and shows increasing amplitude (see 1st row of Figure~\ref{fig:suple-rect}). Although this series includes large shifts, we do not regard those as abrupt changes defined in our paper since rectangular patterns are repeated (i.e., seasonal component). Also, the increasing amplitude (i.e., trend component) is considered one of the trend types. Since we calculate LD by sliding the window, the increasing amplitude does not change the LD values. For example, the LD value of the window, which of size is large enough to cover period, has a value less than 0 (greater than 0 if the amplitude decreases) regardless of time. In this case, since the LD values of all windows are similar, they will be given the same weights. Therefore, even if our method is applied, we would observe more or less the same performance as shown in Rect-Normal dataset of Table~\ref{tab:rect}. 
\begin{table}[]
\caption{Rectangular Time Series with the increasing amplitude and the randomly broken periodicity.}
\resizebox{\textwidth}{!}{
\begin{tabular}{@{}cccc|cc|cc@{}}
\toprule
\multicolumn{2}{c}{Models}                                             & \multicolumn{2}{c}{Pyraformer} & \multicolumn{2}{c}{Autoformer} & \multicolumn{2}{c}{Informer}  \\ 
% \cmidrule{3-18}
% \cmidrule{3-4}

\cmidrule(lr){3-4}
\cmidrule(lr){5-6}
\cmidrule(lr){7-8}

\multicolumn{2}{c}{MSE}                                             & base       & our           & base      & our         & base          & our  \\ \midrule
% ETTh1
\multicolumn{1}{l|}{\multirow{5}{*}{\rotatebox[origin=c]{90}{Rect-\textbf{Normal}}}} & \multicolumn{1}{l|}{96}  & 0.2405 ± \small0.0211 &	0.2468 ± \small0.0243 &	0.9748 ± \small0.4805 &	0.9131 ± \small0.4100 &	0.5409 ± \small0.0280 &	0.5480 ± \small0.0316   \\

\multicolumn{1}{l|}{} & \multicolumn{1}{l|}{168}  &	0.2614 ± \small0.0115 &	0.2652 ± \small0.0102	& 1.4711 ± \small0.9902 & 1.7147 ± \small0.7723 & 1.3573 ± \small0.1377 & 1.3556 ± \small0.1068 \\
\multicolumn{1}{l|}{} & \multicolumn{1}{l|}{336}  & 0.3179 ± \small0.0075 &	0.3193 ± \small0.0062 &	0.5271 ± \small0.2047 &	0.4492 ± \small0.1558 &	1.2945 ± \small0.0874 &	1.2764 ± \small0.0891 \\
\multicolumn{1}{l|}{} & \multicolumn{1}{l|}{720}  & 0.4034 ± \small0.0084 &	0.4088 ± \small0.0089 &	2.7076 ± \small0.4390 &	2.6447 ± \small0.6276 &	1.6796 ± \small0.0450 &	1.7306 ± \small0.0614 \\ \cmidrule(lr){2-8}

\multicolumn{1}{l|}{}                       & \multicolumn{1}{l|}{Imp.} & \multicolumn{2}{c}{1.46\%} & \multicolumn{2}{c}{-1.72\%} & \multicolumn{2}{c}{0.71\%} \\
\midrule

\multicolumn{1}{l|}{\multirow{5}{*}{\rotatebox[origin=c]{90}{Rect-\textbf{Broken}}}} & \multicolumn{1}{l|}{96}  & 0.4028 ± \small0.0434&	0.2343 ± \small0.0061 &	0.9883 ± \small0.1790 &	1.1399 ± \small0.4871 &	0.5781 ± \small0.0558 &	0.3912 ± \small0.0463 \\

\multicolumn{1}{l|}{} & \multicolumn{1}{l|}{168}  &	0.3261 ± \small0.0138 &	0.3020 ± \small0.0174&	1.7361 ± \small0.8130 &	1.2914 ± \small0.3595 &	0.5622 ± \small0.0386&	0.5227 ± \small0.0178 \\
\multicolumn{1}{l|}{} & \multicolumn{1}{l|}{336}  & 0.2548 ± \small0.0112 &	0.2579 ± \small0.0155&	0.7678 ± \small0.1225&	0.5648 ± \small0.1603&	0.5144 ± \small0.0394&	0.4988 ± \small0.0494\\
\multicolumn{1}{l|}{} & \multicolumn{1}{l|}{720}  & 0.3098 ± \small0.0617&	0.3037 ± \small0.0427	& 2.0861 ± \small0.1504 &	1.8670 ± \small0.3713 &	0.5936 ± \small0.0473 &	0.5569 ± \small0.0390 \\ \cmidrule(lr){2-8}

\multicolumn{1}{l|}{}                       & \multicolumn{1}{l|}{Imp.} & \multicolumn{2}{c}{\textbf{-12.49\%}} & \multicolumn{2}{c}{\textbf{-11.80\%}} & \multicolumn{2}{c}{\textbf{-12.14\%}} \\
\bottomrule
\end{tabular}
}
\label{tab:rect}
\end{table}
Additionally, we conducted experiments by removing rectangles randomly from the dataset (see 3rd row of Figure~\ref{fig:suple-rect}). This can be considered abrupt changes (e.g., broken periodicity). We observe that our ReLD brings performance gain in such cases (see Rect-Broken of Table~\ref{tab:rect}). This again demonstrates that our proposed method promotes the model to be robust to abrupt changes.

\subsection{Impact of the in-output ratio}
\begin{table}[]
\caption{Impact of the ratio I/O on multivariate time series forecasting.}
\resizebox{\textwidth}{!}{
\begin{tabular}{@{}cccc|cc|cc||c@{}}
\toprule
\multicolumn{2}{c}{Models}                                             & \multicolumn{2}{c}{Pyraformer} & \multicolumn{2}{c}{Autoformer} & \multicolumn{2}{c}{Informer}   & \\ 
\cmidrule(lr){3-4}
\cmidrule(lr){5-6}
\cmidrule(lr){7-8}

Output-96 & Input                                       & base           & our           & base         & our         & base           & our   & Imp.\\ \midrule
% ETTh1
\multicolumn{1}{l|}{\multirow{5}{*}{\rotatebox[origin=c]{0}{ETTh1}}} & \multicolumn{1}{l|}{48}  & 	0.6314 ± \small0.0371 &	0.5166 ± \small0.0104 &	0.4748 ± \small0.0328 &	0.4675 ± \small0.0504 &	1.0632 ± \small0.2707 &	0.8125 ± \small0.0679 &	-23.58\% \\

\multicolumn{1}{l|}{} & \multicolumn{1}{l|}{96}  &	0.6453 ± \small0.0583 &	0.5345 ± \small0.0073	& 0.4531 ± \small0.0282 & 0.4452 ± \small0.0153 & 0.9075 ± \small0.0479 &	0.8476 ± \small0.0532 &	-6.6\% \\
\multicolumn{1}{l|}{} & \multicolumn{1}{l|}{168}  & 0.6330 ± \small0.0241& 	0.5604 ± \small0.0145 &	0.4477 ± \small0.0247 &	0.4594 ± \small0.0426 &	0.8997 ± \small0.0738 &	0.7928 ± \small0.0698 &	-11.88\% \\
\multicolumn{1}{l|}{} & \multicolumn{1}{l|}{336}  & 0.7195 ± \small0.0206 &	0.6310 ± \small0.0293	& 0.4667 ± \small0.0240 &	0.4826 ± \small0.0188 &	1.1695 ± \small0.2012 &	1.0384 ± \small0.1830 &	-11.21\% \\ 

\multicolumn{1}{l|}{} & \multicolumn{1}{l|}{720}  & 0.7290 ± \small0.0757 &	0.6540 ± \small0.0191 &	0.6354 ± \small0.0386 &	0.5057 ± \small0.0673 &	1.6608 ± \small0.1402 &	1.3866 ± \small0.1341	& -16.51\% \\ 

\midrule
% ETTh2
\multicolumn{1}{l|}{\multirow{5}{*}{\rotatebox[origin=c]{0}{ETTh2}}} & \multicolumn{1}{l|}{48}  & 1.5411 ± \small0.1880 & 1.0982 ± \small0.1702 & 0.3637 ± \small0.0091 &	0.3394 ± \small0.0051	& 1.7225 ± \small0.1508 &1.1461 ± \small0.0743	& -22.96\% \\

\multicolumn{1}{l|}{} & \multicolumn{1}{l|}{96}  &	1.6090 ± \small0.0866 &	1.1733 ± \small0.2271 &	0.3731 ± \small0.0294 & 	0.3464 ± \small0.0102 &	3.4245 ± \small0.4814 &	2.4505 ± \small0.4804 &	-20.89\% \\
\multicolumn{1}{l|}{} & \multicolumn{1}{l|}{168}  & 1.7787 ± \small0.2003 &	1.3081 ± \small0.2461 &	0.4414 ± \small0.0271 &	0.3833 ± \small0.0069 &	5.6370 ± \small0.8005 &	2.9705 ± \small0.4835 	& -28.97\% \\
\multicolumn{1}{l|}{} & \multicolumn{1}{l|}{336}  & 1.7924 ± \small0.2872 &	1.5560 ± \small0.1676 & 0.4897 ± \small0.0565 & 0.4174 ± \small0.0517 &	6.2992 ± \small0.9310 &	3.9496 ± \small0.8050	& -21.75\% \\ 

\multicolumn{1}{l|}{} & \multicolumn{1}{l|}{720}  & 2.0959 ± \small0.1960 &	1.9368 ± \small0.2612	& 0.6769 ± \small0.1552 & 0.4701 ± \small0.0869 & 9.1387 ± \small2.0638 &	6.9792 ± \small1.5690 & -20.59\% \\ 

\midrule
% ETTm1
\multicolumn{1}{l|}{\multirow{5}{*}{\rotatebox[origin=c]{0}{ETTm1}}} & \multicolumn{1}{l|}{48}  & 	0.5559 ± \small0.0225 &	0.4922 ± \small0.0134 &	0.5673 ± \small0.0542 &	0.5182 ± \small0.0396	& 0.6389 ± \small0.0270 &	0.5925 ± \small0.0348 &	-9.13\% \\

\multicolumn{1}{l|}{} & \multicolumn{1}{l|}{96}  &	0.5364 ± \small0.0318 &	0.4713 ± \small0.0299 & 	0.5128 ± \small0.0635 &	0.4545 ± \small0.0410 &	0.6438 ± \small0.0596 &	0.5367 ± \small0.0439 &	-13.38\% \\
\multicolumn{1}{l|}{} & \multicolumn{1}{l|}{168}  &	0.5015 ± \small0.0431 &	0.4174 ± \small0.0176 &	0.4987 ± \small0.0241 &	0.4603 ± \small0.0636 &	0.6907 ± \small0.0578 &	0.5677 ± \small0.0281	& -14.09\% \\
\multicolumn{1}{l|}{} & \multicolumn{1}{l|}{336}  &0.4876 ± \small0.0325 &	0.4316 ± \small0.0171 & 0.5374 ± \small0.0361 &	0.5053 ± \small0.0462 &	0.8487 ± \small0.0578 &	0.6078 ± \small0.0541 
&	-15.28\% \\ 

\multicolumn{1}{l|}{} & \multicolumn{1}{l|}{720}  & 0.4841 ± \small0.0381 &	0.4546 ± \small0.0224 &	0.5799 ± \small0.0914 &	0.4988 ± \small0.0384 &	1.0951 ± \small0.1741 &	0.8007 ± \small0.1602	&-15.65\% \\ 

\bottomrule
\end{tabular}
}
\label{tab:ioimpact}
\end{table}
We conducted an experiment by fixing the output length and changing the input length from 48 to 720 to explore the performance change according to the I/O ratio. We conducted experiments on the three datasets, ETTh1, ETTh2, and ETTm1.
% Applying our method brings consistent performance improvements, although there is an absolute performance difference according to each length.
Applying our method brings consistent performance improvements, although there exists different performance gains depending on the input lengths as shown in Table~\ref{tab:ioimpact}.

\section{Comparison with other methods}
\subsection{Comparison with smoothing and outlier filtering}
\label{apx:smod}
We compared our proposed method with 1) smoothing and 2) outlier filtering which are expected to perform well with drastic changes (e.g., fluke) in time-series datasets. Smoothing techniques are used to remove nosiness and reduce outliers, allowing meaningful temporal patterns to stand out. Conventional methods include moving average (MA) smoothing as follows:
\begin{equation}
    s_{t}=\frac{\left(x_{t-k+1}+x_{t-k+2}+ \ldots+x_{t}\right)}{k}
\end{equation} where $s_t$ is the smoothed observation at $t$ and $x_t$ is the original observation.
The other method is exponential (EMA) smoothing calculated by Equation as follows:
\begin{equation}
    s_{t}=\alpha \cdot x_{t}+(1-\alpha) \cdot s_{t-1}
\end{equation} where $\alpha \in (0, 1)$. 
We smoothed the training time series and train forecasting models. To use outlier filtering method for forecasting task, a simple way to detect outliers is to assume that the target value follows a Gaussian and remove values that exceed a certain range of values. 
% we remove outliers over a certain value and training forecasters.
We train forecasters after removing outliers which exceed a certain value.

\subsection{Comparison with error-based reweighting}
\label{apx:error-re}
As we mentioned in the main paper, we observed that abrupt changes significantly contribute to the total loss in the training phase. 
In this situation, we can simply reweight a loss of sample that have large error while considering the sample including abrupt change. 
Reweighting inversely to the error may down-weight the loss of the abrupt change without additional LD calculation. In the main paper, we presented two error-based methods: Focal-R and filp Focal-R. Focal-R loss is calculated as $\sigma\left(\beta\left| e_i\right|\right)^\gamma L_i$ where $e_i$ is error of $i$-th sample, $L_i$ is loss of $i$-th sample, and $\sigma(\cdot)$ is sigmoid function. $\beta$ and $\gamma$ are hyperparaters. In case of filp Focal-R, $\beta$ is negative to flip the sigmoid function along the $y$ axis.
Additionally, we provide L2 error-based reweighting results, namely invL2 which is written as $\frac{L_i}{e_i + \epsilon}$. In case of invL2, as the model forecasts accurately and thus the error of the normal states is close to zero, the parameter moves with larger steps by up-weighted loss. Table~\ref{tab:loss_based} shows the performance in the case of reweighting inversely to the error of each window.

\begin{table}[]
% \caption{Comparisons with re-weighting \js{based approaches} on loss in multivariate forecasting (Top) and in univariate forecasting (Bottom) on ETTh1 dataset.}
\caption{Comparison with error-based reweighting (invL2) in the multivariate forecasting (Top) and in the univariate forecasting (Bottom) using ETTh1 dataset.}
\resizebox{\textwidth}{!}{
\begin{tabular}{c|ccc|ccc|ccc}
\toprule
Multivariate     & \multicolumn{3}{c}{Pyraformer}             & \multicolumn{3}{c}{Autoformer}             & \multicolumn{3}{c}{Informer}                  \\  

\cmidrule(lr){2-4}
\cmidrule(lr){5-7}
\cmidrule(lr){8-10}

I / O & base &invL2  & \multicolumn{1}{c|}{ReLD} & base &invL2 & \multicolumn{1}{c|}{ReLD} & base &invL2  & \multicolumn{1}{c|}{ReLD}         \\ \midrule

96 / 96 & 0.6453 ± \small0.0583 &	0.6083 ± \small0.0149&	0.5345 ± \small0.0073 &	0.4422 ± \small0.0242 &	0.4458 ± \small0.0212 &	0.4438 ± \small0.0143 &	0.9084 ± \small0.0485 &	0.8506 ± \small0.0280 &	0.8031 ± \small0.0317 \\
336 / 168 &	0.8644 ± \small0.0905&	0.7842 ± \small0.0250 & 0.7415 ± \small0.0399 &	0.5042 ± \small0.0515 &	0.4772 ± \small0.0144 &	0.4906 ± \small0.0263 &	1.3720 ±  \small0.2422 &	1.2150 ± \small0.1333 &	0.8858 ± \small0.0258 \\
336 / 336 &	0.9328 ± \small0.0341 &	0.9643 ± \small0.0404 &	0.8895 ± \small0.0548 &	0.5694 ± \small0.1115 &	0.5450 ± \small0.0886 &	0.5110 ± \small0.0990 &	1.3425 ± \small0.0725 &	1.2857 ± \small0.0710 &	0.9850 ± \small0.0308 \\
336 / 720 & 0.9843 ± \small0.0213 &	1.0003 ± \small0.0228 &	0.9781 ± \small0.0196 &	0.5348 ± \small0.0212 &	0.5589 ± \small0.0613 &	0.5207 ± \small0.0106 &	1.3933 ± \small0.0892 &	1.3735 ± \small0.0386 &	1.1994 ± \small0.0597 \\
\midrule
Imp. &	- &	-2.50\% &	\textbf{-9.17\%}	 &-	 & -1.08\%	& \textbf{-3.81\%}	& -	& -5.86\%	& \textbf{-21.89\%} \\

\bottomrule
\end{tabular}
}
\vspace{3mm}

\resizebox{\textwidth}{!}{
\begin{tabular}{c|ccc|ccc|ccc}
\toprule
Univariate     & \multicolumn{3}{c}{Pyraformer}             & \multicolumn{3}{c}{Autoformer}             & \multicolumn{3}{c}{Informer}                  \\  

\cmidrule(lr){2-4}
\cmidrule(lr){5-7}
\cmidrule(lr){8-10}

I / O & base &invL2  & \multicolumn{1}{c|}{ReLD} & base &invL2  & \multicolumn{1}{c|}{ReLD} & base &invL2  & \multicolumn{1}{c|}{ReLD}         \\ \midrule

96 / 96 & 0.2074 ± \small0.0728 &	0.1928 ± \small0.0365 &	0.1831 ± \small0.0533 &	0.0859 ± \small0.0063 &	0.0861 ± \small0.0031 &	0.0841 ± \small0.0067 &	0.1203 ± \small0.0730 &	0.1132 ± \small0.0441 &	0.1020 ± \small0.0472 \\
336 / 168 &	0.1819 ± \small0.0257 &	0.1750 ± \small0.0581 &	0.1725 ± \small0.0406 &	0.1077 ± \small0.0130 &	0.0949 ± \small0.0109 &	0.0999 ± \small0.0072 &	0.0862 ± \small0.0292 &	0.0946 ± \small0.0244 &	0.0848 ± \small0.0297 \\
336 / 336 &	0.1716 ± \small0.0597 &	0.1853 ± \small0.0622 &	0.1649 ± \small0.0426 &	0.1055 ± \small0.0219 &	0.1135 ± \small0.0198 &	0.1008 ± \small0.0157 &	0.0862 ± \small0.0025 &	0.0870 ± \small0.0084 & 	0.0897 ± \small0.0165 \\
336 / 720 &0.1974 ± \small0.0415 &	0.1746 ± \small0.0312 &	0.1667 ± \small0.0298 &	0.1352 ± \small0.0207 &	0.1251 ± \small0.0110 &	0.1244 ± \small0.0270 &	0.2025 ± \small0.0961 &	0.1800 ± \small0.0945 &	 0.1550 ± \small0.0237 \\
\midrule
Imp. & - &	-3.60\%	& \textbf{-9.09\%} & -	& -2.88\% & \textbf{-5.45\%} &	- &	-1.59\%	& \textbf{-9.06\%} \\

\bottomrule
\end{tabular}
}

\label{tab:loss_based}
\end{table}

\subsection{Variants for Local Discrepancy}
\label{apx:variousld}
We propose the \textit{Local Discrepancy} (LD) based on the statistics formulated by a statistical test, Welch's t-test~\cite{welch1938significance}, in order to measure how two adjacent in-output sequences, $\mathcal{X}_t$ and $\mathcal{Y}_t$, are different from each other.
There may exist other metrics to measure the local discrepancy such as multivariate t-statistic~\cite{hotelling1992generalization} (i.e., Hotelling's $t$-squared statistic) and stationarity tests (\textit{e.g.}, Kwiatkowski–Phillips–Schmidt–Shin (KPSS) tests~\cite{kwiatkowski1992testing}).
We also report the performance of our reweighting framework using a different metric other than t-statistics for measuring the local discrepancy in Table~\ref{tab:stats}. 
Hotelling's $t$-squared statistic is a generalization of Student's $t$-statistic that is used in multivariate hypothesis testing. We can naturally utilize $t$-squared statistic as LD for multivariate forecasting (\textit{i.e.}, $\mathbf{s} \in \mathbb{R}^{m}$ and $ m > 1$) as follows:
\begin{equation}
    \operatorname{LocalDis}(\mathcal{X}_t, \mathcal{Y}_t)=\frac{I\cdot O}{I + O}(\mathcal{\bar{X}}_t-\mathcal{\bar{Y}}_t)^{\prime} \hat{\boldsymbol{\Sigma}}^{-1}(\mathcal{\bar{X}}_t-\mathcal{\bar{Y}}_t) \coloneqq v^2_t
\end{equation}
where the mean and covariance are defined as follows: 
\begin{equation*}
\mathcal{\bar{X}}_t=\frac{1}{I} \sum_{i=1}^{I} \mathbf{s}_{t-i}, \quad \mathcal{\bar{Y}}_t=\frac{1}{O} \sum_{i=0}^{O-1} \mathbf{s}_{t+i}, \quad
\hat{\boldsymbol{\Sigma}}=\frac{\left(I-1\right) \hat{\boldsymbol{\Sigma}}_{\mathcal{\bar{X}}}+\left(O-1\right) \hat{\boldsymbol{\Sigma}}_{\mathcal{\bar{Y}}}}{I+O-2},
\end{equation*}
\begin{equation*}
\hat{\boldsymbol{\Sigma}}_{\mathcal{\bar{X}}}=\frac{1}{I-1} \sum_{i=1}^{I }\left(\mathbf{s}_{t-i}-\mathcal{\bar{X}}_t\right)\left(\mathbf{s}_{t-i}-\mathcal{\bar{X}}_t\right)^{\prime}, \quad
\hat{\boldsymbol{\Sigma}}_{\mathcal{\bar{Y}}}=\frac{1}{O-1} \sum_{i=1}^{O-1}\left(\mathbf{s}_{t+i}-\mathcal{\bar{Y}}\right)\left(\mathbf{s}_{t+i}-\mathcal{\bar{Y}}\right)^{\prime}.
\end{equation*}
We can interpret the time-series data in terms of stochastic processes. KPSS tests are used for testing a null hypothesis that an observable time series is stationary around a deterministic trend (\textit{i.e.}, trend-stationary) against the alternative of a unit root. When the given time series is trend stationary, the KPSS statistic has small value, which is close to zero. Thus, to measure the degree of abruptness of a change in a given period of time, we leverage the KPSS statistic as LD:

\begin{equation}
\operatorname{LocalDis}(\operatorname{concat}(\mathcal{X}_t, \mathcal{Y}_t)) = \frac{1}{(I+O)^{2}} \cdot \sum_{i=-I}^{O-1} \frac{\mathcal{E}_{t+i}^{2}}{\hat{\sigma}^{2}} \coloneqq v_t
\end{equation} where $\mathcal{E}_t$ is partial sum of the residuals and $\hat{\sigma}^{2}$ is the estimate of the long-run variance of the residuals as follows: 
\begin{equation*}
\mathcal{E}_{k}=\sum_{k=1}^{t} e_{i}, \quad e = (e_{t-I}, e_{t-I+1}, \dots, e_{O-1})
\end{equation*} where $e$ means OLS residuals when regressing the concated in-output sequence (\textit{i.e.}, $\operatorname{concat}(\mathcal{X}_t, \mathcal{Y}_t)$). 
\begin{table}[]
% \caption{
% Ablation study on variants \js{of} local discrepancy, used in our reweighting framework.
% We compare models 1) KPSS, 2) $t$-squared, and 3) $t$-statistic. The $t$-statistic shows more consistent and superior results compared to other statistics in the multivariate setting.
% }
\caption{
Ablation study on variants of local discrepancy used in our reweighting framework.
We compare models which uses 1) KPSS, 2) $t$-squared, and 3) $t$-statistic. The $t$-statistic shows more consistent and superior results compared to other statistics in the multivariate setting.}
\resizebox{\textwidth}{!}{
\begin{tabular}{lc|ccc|ccc|ccc||r}
\toprule
 & Dataset                         & \multicolumn{3}{c}{ETTh1} & \multicolumn{3}{c}{ETTh2} & \multicolumn{3}{c||}{ETTm1} \\

\cmidrule(lr){3-5}
\cmidrule(lr){6-8}
\cmidrule(lr){9-11}
% \cmidrule(lr){9}

       Model & Predict-O                          & 96     & 168     & 336  & 96     & 168     & 336    & 96     & 168     & 336    &   Imp.    \\ \midrule
\multicolumn{2}{l|}{Pyraformer}                        & 0.645 & 0.864 & 0.933 & 1.609 & 5.014 & 4.356 & 0.536 & 0.563 & 0.697 &  -     \\
\multicolumn{2}{l|}{Pyraformer + KPSS}                 &  0.554 & 0.782 & 0.909 & 1.482 & 4.590 & 5.327 & 0.470 & 0.527 & 0.604 &     -5.84\%  \\
\multicolumn{2}{l|}{Pyraformer + $t$-squared} &  0.640 & 0.809 & 0.898 & 1.440 & 3.112 & 3.912 & 0.490 & 0.557 & 0.632  &  -9.84\%    \\
\multicolumn{2}{l|}{Pyraformer + $t$-statistic}                    & 0.534 & 0.742 & 0.889 & 1.173 & 3.976 & 3.281 & 0.471 & 0.506 & 0.573  &  \textbf{-16.51\%}    \\ \midrule
\multicolumn{2}{l|}{Autoformer}                        & 0.442 & 0.504 & 0.569 & 0.386 & 0.439 & 0.494 & 0.524 & 0.534 & 0.561 &   -   \\ 
\multicolumn{2}{l|}{Autoformer + KPSS}                 & 0.446 & 0.528 & 0.486 & 0.358 & 0.436 & 0.516 & 0.456 & 0.538 & 0.513   &  -3.68\%    \\
\multicolumn{2}{l|}{Autoformer + $t$-squared} & 0.454 & 0.521 & 0.515 & 0.357 & 0.403 & 0.436 & 0.503 & 0.548 & 0.512  &  -4.60\%    \\
\multicolumn{2}{l|}{Autoformer + $t$-statistic}                    & 0.444 & 0.491 & 0.511 & 0.351 & 0.413 & 0.424 & 0.455 & 0.500 & 0.514 & \textbf{-7.74\%}  \\ \midrule
\multicolumn{2}{l|}{Informer}                          & 0.908 & 1.372 & 1.343 & 3.400 & 5.796 & 3.901 & 0.640 & 1.224 & 1.390 
 &  -    \\
\multicolumn{2}{l|}{Informer + KPSS}                  &  0.850 & 1.215 & 1.215 & 3.050 & 5.593 & 4.202 & 0.535 & 0.844 & 1.087  &  -11.41\%    \\
\multicolumn{2}{l|}{Informer + $t$-squared}   &  0.871 & 1.262 & 1.234 & 2.796 & 4.393 & 3.419 & 0.594 & 0.992 & 1.195  & -12.76\%     \\
\multicolumn{2}{l|}{Informer + $t$-statistic}                      &  0.856 & 1.113 & 1.151 & 2.462 & 4.723 & 3.788 & 0.543 & 0.751 & 1.008  &  \textbf{-18.81\%}
\\
\bottomrule
\end{tabular}
}
\label{tab:stats}
\end{table}
% As shown in Table~\ref{tab:stats}, we compared the both $t$-Squared and KPSS with the baselines by replacing $t$-statistic in our ReLD framework. 
% When leveraging the both $t$-Squared and KPSS as LD, the MSE of three recent models with ReLD was consistently lower than the baselines as when using $t$-statistic. 
% This means that our reweighting framework is useful as it improves performance without relying on specific statistic formula. Moreover, we empirically confirmed that t-statistic is more suitable as LD, obtaining the more performance gain compared to the other two statistics. 
We observe that our reweighting framework consistently outperforms the ones without our framework regardless of the statistics used for measuring the local discrepancy. 
While we empirically confirmed that using t-statistic is more suitable for LD compared to KPSS or t-Squared statistic, such result demonstrates that our framework can be used with any statistics measure the user deems appropriate.

\section{Our Framework Details}
\label{apx:framework}

\subsection{Implementation details}
We include 12 baselines to validate our ReLD. All models were implemented with PyTorch. As for recent models  (\textit{i.e.}, FEDformer\footnote{https://github.com/MAZiqing/FEDformer}, Pyraformer\footnote{https://github.com/alipay/Pyraformer}, Autoformer\footnote{https://github.com/thuml/Autoformer}, and Informer\footnote{https://github.com/zhouhaoyi/Informer2020}), we used the official code released by the original authors, rather than implementing from scratch. For a fair comparison between ReLD and the existing framework, we set the same hyperparameters found in each work. We trained all models from scratch to 10 epochs.
%or all experiments. 
To assign weights to all training samples in ReLD, the LD is computed only once before training, and it takes only a negligible amount of time compared to the training time. Most models, which leverage a generative decoding, take an average of less than an hour to train on a TITAN-Xp GPU except for LSTMa which uses auto-regressive decoding.

\subsection{Dataset details}
In this work, we reported the results on eight datasets. ETT are widely used to evaluate long-term forecasting methods~\cite{zhou2021informer,wu2021autoformer,liu2021pyraformer,zhou2022fedformer} 
ETT contains the crucial indicators (e.g., oil temperature, load, etc) collected from the electricity transformers over two years, and are categorized into four datasets depending on the location (ETT1 and ETT2) and interval (15 minutes and one hour). 
Electricity~\footnote{https://archive.ics.uci.edu/ml/datasets/ElectricityLoadDiagrams20112014} dataset
contains the hourly electricity consumption of 321 customers from 2012 to 2014. 
Weather dataset~\footnote{https://www.bgc-jena.mpg.de/wetter/} is recorded every 10 minutes for a year, which contains 21 meteorological indicators (\textit{e.g.}, air temperature, humidity, etc). 
Pump dataset~\footnote{https://www.kaggle.com/datasets/nphantawee/pump-sensor-data} is collected from 52 sensors monitoring the water pump. 
AirQuality dataset~\footnote{https://archive.ics.uci.edu/ml/datasets/air+quality}, taken from the UCI repository, contains hourly averaged responses obtained from five metal oxide chemical sensors of an air quality chemical multi-sensor device.

\subsection{Pseudo code for ReLD}
\begin{algorithm}
\caption{\textbf{ReLD}: \textbf{Re}weighting framework based on \textbf{L}ocal Discrepancy \textbf{D}ensity}\label{alg:cap}
\begin{algorithmic}
\Require Training set $\mathcal{D}=\left\{\left(\mathcal{X}_{t}, \mathcal{Y}_{t}\right)\right\}_{t=1}^{N}$, bin size $\Delta b$, symmetric kernel distribution $k(v, v')$\\
Compute Local Discrepancy $\operatorname{LD}\left(\mathcal{X}_{t}, \mathcal{Y}_{t}\right)=\frac{\bar{\mathcal{X}}_{t}-\bar{\mathcal{Y}}_{t}}{\sqrt{\frac{s_{\bar{\mathcal{X}}_{t}}^{2}}{I}+\frac{s_{\bar{\mathcal{Y}}_{t}}^{2}}{O}+\varepsilon}} \coloneqq v_t$\\
Compute the empirical label density distribution $p(v)$ based on $\Delta b$ and $\mathcal{D}$ \\
Compute the effective label density distribution $\tilde{p}\left(v^{\prime}\right) \coloneqq \int_{\mathcal{V}} k\left(v, v^{\prime}\right) p(v) d v$

\ForAll{$\left(\mathcal{X}_{t}, \mathcal{Y}_{t},  v_t\right) \in \mathcal{D}$}
\State Assign weight for each sample as $w_{t} \propto c \cdot \tilde{p}\left(v_{t}\right)$ (constant $c$ as scaling factor)
\EndFor

\ForAll{number of training iterations}
\State Sample a mini-batch $\left\{\left(\mathcal{X}_{b},  \mathcal{Y}_{b}, w_{b}\right)\right\}_{b=1}^{B}$ from $\mathcal{D}$
\State Forward $\left\{\mathcal{X}_{b}\right\}_{b=1}^{B}$ and get corresponding predictions $\{\hat{\mathcal{Y}}_{b}\}_{b=1}^{B}$
\State Do one training step using the weighted loss $\frac{1}{B} \sum_{b=1}^{B} \mathcal{L}_{w_b}(\hat{\mathcal{Y}}_{b}, \mathcal{Y}_{b})$
\EndFor

\end{algorithmic}
\label{algo:reld}
\end{algorithm}

% ReLD: Re-weighting based on Local discrepancy Density 

We illustrate the pseudo code of the ReLD in Algorithm~\ref{algo:reld}.

\subsection{Hyperparameter Sensitivity}
\begin{table}[]
\caption{
Performance change according to the number of bins.
}
\centering

\begin{tabular}{l|c|c|c||r}
\toprule
 \multicolumn{2}{c|}{Dataset}                          & ETTh1 & ETTh2 &  \\
\midrule
% \cmidrule(lr){3-5}
% \cmidrule(lr){6-8}
% \cmidrule(lr){9-11}
% \cmidrule(lr){9}

       Model &  \# bins                          &  336  →  336  & 96 →  96    &   Imp.    \\ \midrule \midrule
Autoformer       &          -       & 0.5694 ± \small0.1115 & 0.3859 ± \small0.0260 &  -     \\
Autoformer + ReLD  &        40        &  0.5245 ± \small0.1543 &	0.3529 ± \small0.0262  &     -8.55\%  \\
 & 120 &0.4907 ± \small0.0337 &	0.3455 ± \small0.0229 &  -10.47\%    \\
 & 200 &0.4903 ± \small0.0610 &	0.3501 ± \small0.0168 &   -9.28\%    \\
 & 300 &0.4881 ± \small0.0413 &	0.3472 ± \small0.0072 &   -10.03\%    \\ 
 & 500 &0.5130 ± \small0.0529 &	0.3485 ± \small0.0142  &  -9.69\%    \\
\bottomrule
\end{tabular}

\vspace{5mm}
\caption{
Performance change according to the KDE kernel types.
}
\begin{tabular}{l|c|c|c||r}
\toprule
 \multicolumn{2}{c|}{Dataset}                          & ETTh1 & ETTh2 &  \\
\midrule
% \cmidrule(lr){3-5}
% \cmidrule(lr){6-8}
% \cmidrule(lr){9-11}
% \cmidrule(lr){9}

       Model &  KDE kernel                          &  336  →  336  & 96 →  96    &   Imp.    \\ \midrule \midrule
Autoformer       &          -       & 0.5694 ± \small0.1115 & 0.3859 ± \small0.0260 &  -     \\
Autoformer + ReLD  &        Gaussian        & 0.4903 ± \small0.0610 &	0.3501 ± \small0.0168  &     -9.28\%  \\
 & Triangle &0.4792 ± \small0.0171 &	0.3453 ± \small0.0232 &   -10.52\%    \\
 & Laplace &0.4786 ± \small0.0380 &	0.3496 ± \small0.0087 &   -9.41\%    \\
\bottomrule
\end{tabular}

\vspace{5mm}
\caption{
Performance changes according to the KDE kernel size.
}
\begin{tabular}{l|c|c|c||r}
\toprule
 \multicolumn{2}{c|}{Dataset}                          & ETTh1 & ETTh2 &  \\
\midrule
% \cmidrule(lr){3-5}
% \cmidrule(lr){6-8}
% \cmidrule(lr){9-11}
% \cmidrule(lr){9}

       Model &  KDE kernel size                          &  336  →  336  & 96 →  96    &   Imp.    \\ \midrule \midrule
Autoformer       &          -       & 0.5694 ± \small0.1115 & 0.3859 ± \small0.0260 &  -     \\
Autoformer + ReLD  &        5        & 0.4903 ± \small0.0610 &	0.3501 ± \small0.0168  &    -9.28\%  \\
 & 10 &0.4896 ± \small0.0728 &	0.3466 ± \small0.0018 &   -10.18\%   \\
 & 15 &	0.4836 ± \small0.0189 &	0.3567 ± \small0.0293 &   -7.57\%    \\
 & 20 & 0.4841 ± \small0.0444 &	0.3482 ± \small0.0383 &   -9.77\%    \\ 
 & 25 &	0.4835 ± \small0.0317 &	0.3490 ± \small0.0166  &   -9.56\%    \\
\bottomrule
\end{tabular}

\vspace{5mm}
\caption{
Performance change according to the KDE kernel sigma.
}
\begin{tabular}{l|c|c|c||r}
\toprule
 \multicolumn{2}{c|}{Dataset}                          & ETTh1 & ETTh2 &  \\
\midrule
% \cmidrule(lr){3-5}
% \cmidrule(lr){6-8}
% \cmidrule(lr){9-11}
% \cmidrule(lr){9}

       Model &  KDE kernel sigma                          &  336  →  336  & 96 →  96    &   Imp.    \\ \midrule \midrule
Autoformer       &          -       & 0.5694 ± \small0.1115 & 0.3859 ± \small0.0260 &  -     \\
Autoformer + ReLD  &        1        & 0.5545 ± \small0.1916 &	0.3521 ± \small0.0189  &     -8.76\%  \\
 & 2 &0.4903 ± \small0.0610 &	0.3501 ± \small0.0168 &  -9.28\%    \\
 & 4 &	0.5218 ± \small0.1632 &	0.3586 ± \small0.0567 &   -7.07\%    \\
 & 8 & 0.4767 ± \small0.0307 &		0.3435 ± \small0.0172 &  -10.99\%    \\ 
 & 16 &	0.4836 ± \small0.0318 & 0.3494 ± \small0.0310 &   -9.46\%    \\
\bottomrule
\end{tabular}

\label{tab:kde_param}
\end{table}

We used KDE to smooth the LD distribution. Related parameters include the bin size that determines how many sections continuous LD is divided into, KDE's kernel type, kernel size and kernel sigma. In our experiment, we set the bin size to 200, kernel type to Gaussian, and kernel size and sigma to 5 and 2, respectively, as default parameters. We conducted experiments on ETTh1 and ETTh2 to observe the variance of performance according to each parameter.
% Although there is a slight performance difference across different parameters, the performance improvements were stable and significant.
As shown in Table~11, Table~12, Table~13, and Table~14, we observe that our proposed method is robust to the hyper-parameters while showing consistent performance improvements.

\begin{figure}[h!]
\begin{center}
  \includegraphics[width=0.85\linewidth]{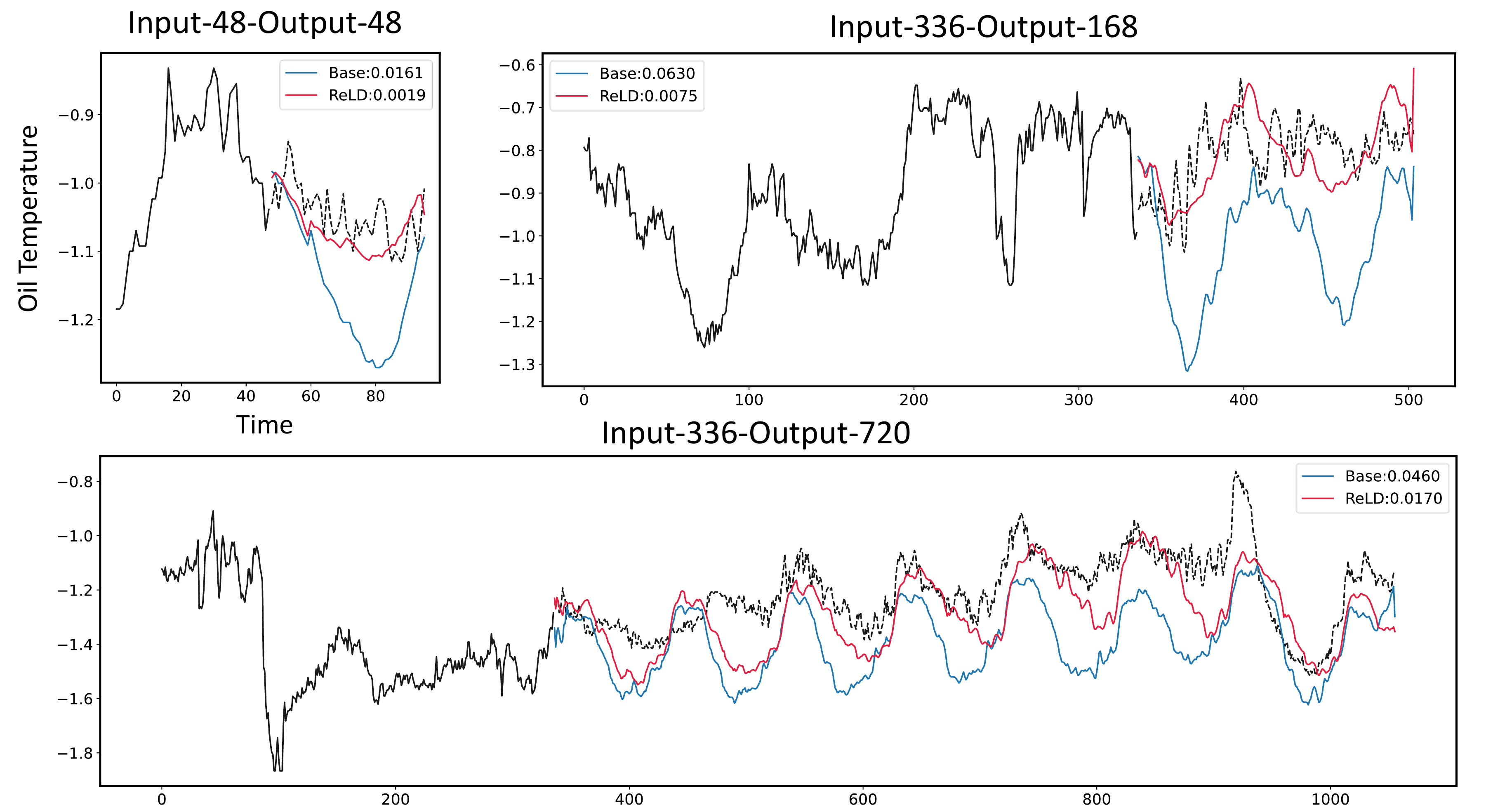}
\end{center}
\vspace{-5mm}
\caption{Forecasting results of Autoformer trained on ETTm1 with three different length settings: Input-48-Output-48, Input-336-Output-168, and Input-336-Output-720. 
The blue line indicates the forecasting results of the baselines without our ReLD and the red line indicates those with our ReLD.
}
\label{fig:suple-length}
\end{figure}
\begin{figure}[h!]
\begin{center}
  \includegraphics[width=0.85\linewidth]{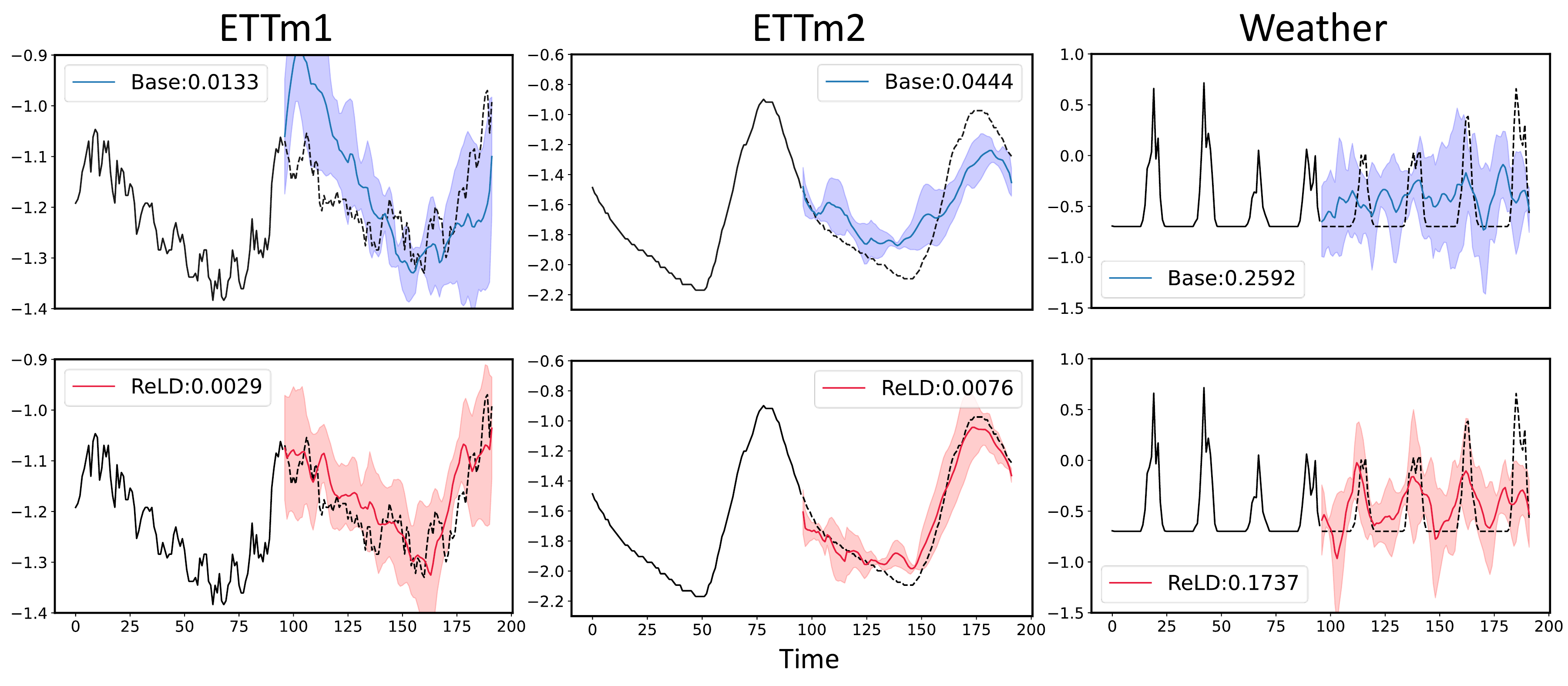}
\end{center}
\vspace{-5mm}
\caption{Forecasting results of Autoformer on three datasets: ETTm1, ETTm2, and Weather. The first row shows the forecasting results of the baseline without our ReLD and the second row shows those with our ReLD.}
\label{fig:suple-dataset}
\end{figure}
\begin{figure}[h!]
\begin{center}
  \includegraphics[width=0.85\linewidth]{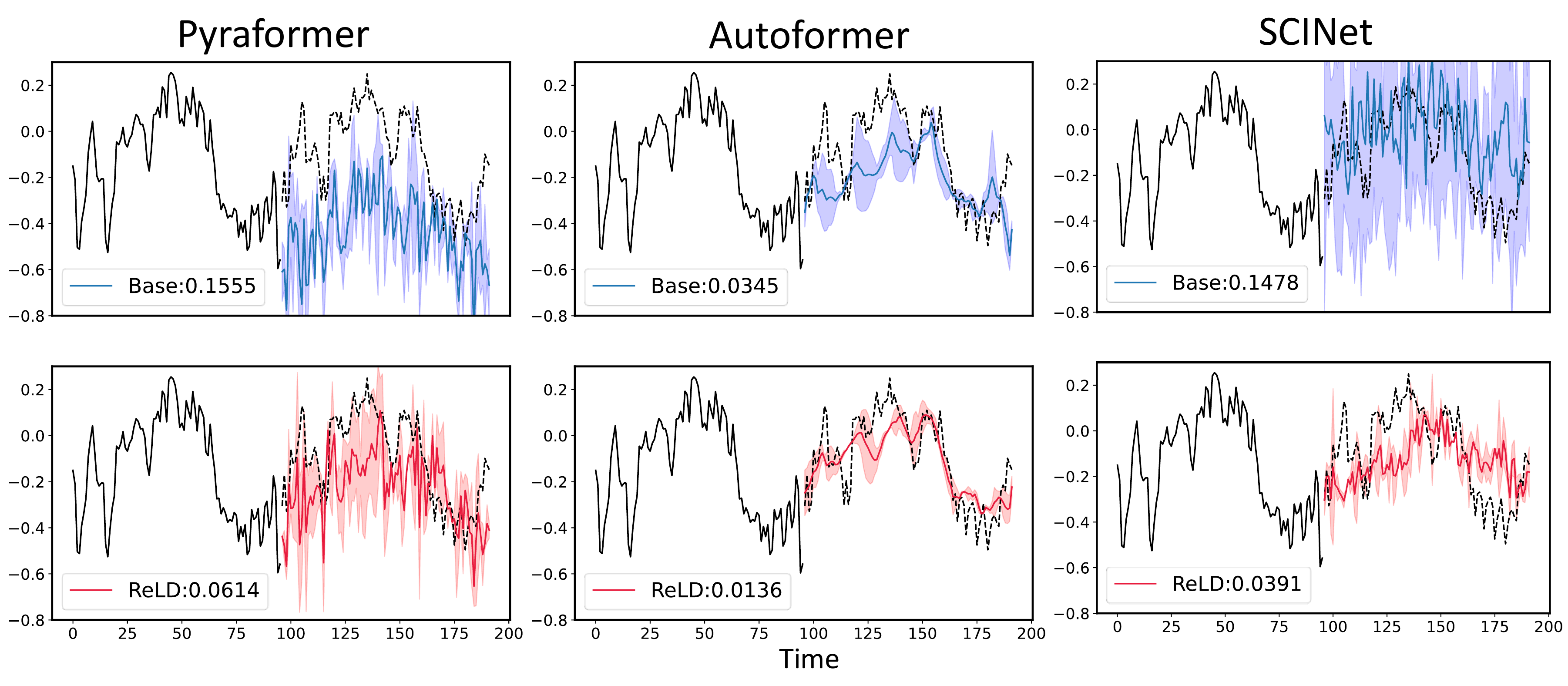}
\end{center}
\vspace{-5mm}
\caption{Forecasting results of the recent three models on the same sample in HULL series of ETTm2. The first row shows the forecasting results of the baselines without our ReLD and the second row shows those with our ReLD.}
\label{fig:suple-model}
\end{figure}

\section{Qualitative Results}
\label{apx:quali}
This section visualizes the forecasting results using three criteria: in-output length (Figure~\ref{fig:suple-length}), dataset (Figure~\ref{fig:suple-dataset}), and model architecture (Figure~\ref{fig:suple-model}). All samples are from the test set of each dataset.
The solid black line denotes the input series and the dotted black line denotes the ground truth series that a model should predict. For a reliable comparison, we plot the averaged forecasting results of the independent models trained from different random initializations. 
% The shaded part of the forecasting result means the point-wise variance.
The shaded part of the forecasting result indicates the forecasting variation at a given time stamp. In Figure~\ref{fig:suple-length}, we only report the mean of forecasting results without the forecasting variation for better clarity. 

As shown in Figure~\ref{fig:suple-length}, our ReLD demonstrated enhanced forecasting results in both short-term and long-term settings. 
% We also confirmed that our ReLD showed more accurate forecasting results for the normal states, which are typical temporal patterns shown in the three datasets (see Figure~\ref{fig:suple-dataset}). 
% For the same sample shown in Figure~\ref{fig:suple-model}, the three recent model has different errors, but when ReLD was applied, the predicted mean and variance overlapped with the ground truth, resulting in the lower MSE than their original results.
We observe that applying ReLD significantly reduces the MSE loss regardless of datasets (see Figure~\ref{fig:suple-dataset}) and model architectures (Figure~\ref{fig:suple-model}).
For example, by applying ReLD on Weather dataset (see Figure~\ref{fig:suple-dataset}), the prediction variations (red-shaded regions) are fitted to the fluctuations of the target times series which was underfitted without applying ReLD (blue-shaded regions). 

\section{Full Benchmark on the Real-World Datasets}

\label{apx:fullbench}
\begin{sidewaystable}[]
\centering
\footnotesize
\resizebox{1.05\textheight}{!}{
\begin{tabular}{@{}cccccccccccccccccccccc@{}}
\toprule
\multicolumn{2}{c}{Models}                                             & \multicolumn{2}{c}{FEDformer} & \multicolumn{2}{c}{Pyraformer} & \multicolumn{2}{c}{Autoformer}  & \multicolumn{2}{c}{Informer} & \multicolumn{2}{c}{LogTrans} & \multicolumn{2}{c}{Reformer} & \multicolumn{2}{c}{Transformer} & \multicolumn{2}{c}{LSTNet} & \multicolumn{2}{c}{LSTMa} & \multicolumn{2}{c}{TCN} \\ \midrule
\multicolumn{2}{c}{MSE}                                             & base           & our           & base         & our         & base           & our           & base          & our         & base          & our          & base          & our          & base          & our          & base            & our           & base         & our         & base         & our       \\ \midrule

% ETTh1
\multicolumn{1}{l|}{} & \multicolumn{1}{l|}{48}  & \makecell{ \textbf{0.3479} \\\scriptsize$\pm$0.0017 } & \makecell{ 0.3483 \\\scriptsize$\pm$0.0017 } & \makecell{ 0.5354 \\\scriptsize$\pm$0.0157 } & \makecell{ 0.4438 \\\scriptsize$\pm$0.0139 } & \makecell{ 0.4160 \\\scriptsize$\pm$0.0216 } & \makecell{ 0.3842 \\\scriptsize$\pm$0.0181 } & \makecell{ 0.6872 \\\scriptsize$\pm$0.0736 } & \makecell{ 0.5734 \\\scriptsize$\pm$0.0409 } & \makecell{ 0.7247 \\\scriptsize$\pm$0.0764 } & \makecell{ 0.5902 \\\scriptsize$\pm$0.0473 } & \makecell{ 0.7393 \\\scriptsize$\pm$0.0260 } & \makecell{ 0.6409 \\\scriptsize$\pm$0.0173 } & \makecell{ 0.6615 \\\scriptsize$\pm$0.0767 } & \makecell{ 0.5443 \\\scriptsize$\pm$0.0370 } & \makecell{ 0.6272 \\\scriptsize$\pm$0.0115 } & \makecell{ 0.5586 \\\scriptsize$\pm$0.0069 } & \makecell{ 0.6938 \\\scriptsize$\pm$0.0447 } & \makecell{ 0.5809 \\\scriptsize$\pm$0.0334 } & \makecell{ 0.5655 \\\scriptsize$\pm$0.0287 } & \makecell{ 0.4862 \\\scriptsize$\pm$0.0151 }  \\
\bottomrule
\end{tabular}
}
\end{sidewaystable}
\begin{sidewaystable}[]
\centering
\footnotesize
\resizebox{1.00\textheight}{!}{
\begin{tabular}{@{}cccccccccccccccccccccc@{}}
\toprule
\multicolumn{2}{c}{Models}                                             & \multicolumn{2}{c}{FEDformer} & \multicolumn{2}{c}{Pyraformer} & \multicolumn{2}{c}{Autoformer}  & \multicolumn{2}{c}{Informer} & \multicolumn{2}{c}{LogTrans} & \multicolumn{2}{c}{Reformer} & \multicolumn{2}{c}{Transformer} & \multicolumn{2}{c}{LSTNet} & \multicolumn{2}{c}{LSTMa} & \multicolumn{2}{c}{TCN} \\ \midrule
\multicolumn{2}{c}{MSE}                                             & base           & our           & base         & our         & base           & our           & base          & our         & base          & our          & base          & our          & base          & our          & base            & our           & base         & our         & base         & our       \\ \midrule

% Pump
\multicolumn{1}{l|}{} & \multicolumn{1}{l|}{48}  & \makecell{ 0.5246 \\\scriptsize$\pm$0.0024 } & \makecell{ \textbf{0.5227} \\\scriptsize$\pm$0.0012 } & \makecell{ 0.7242 \\\scriptsize$\pm$0.0184 } & \makecell{ 0.7151 \\\scriptsize$\pm$0.0153 } & \makecell{ 0.5767 \\\scriptsize$\pm$0.0364 } & \makecell{ 0.5439 \\\scriptsize$\pm$0.0042 } & \makecell{ 0.8648 \\\scriptsize$\pm$0.0294 } & \makecell{ 0.7934 \\\scriptsize$\pm$0.1065 } & \makecell{ 0.7886 \\\scriptsize$\pm$0.0569 } & \makecell{ 0.7129 \\\scriptsize$\pm$0.0673 } & \makecell{ 0.8083 \\\scriptsize$\pm$0.0318 } & \makecell{ 0.7282 \\\scriptsize$\pm$0.0155 } & \makecell{ 0.8434 \\\scriptsize$\pm$0.0664 } & \makecell{ 0.6645 \\\scriptsize$\pm$0.0697 } & \makecell{ 0.8234 \\\scriptsize$\pm$0.0376 } & \makecell{ 0.7343 \\\scriptsize$\pm$0.0110 } & \makecell{ 0.7540 \\\scriptsize$\pm$0.0238 } & \makecell{ 0.7155 \\\scriptsize$\pm$0.0295 } & \makecell{ 0.8649 \\\scriptsize$\pm$0.0920 } & \makecell{ 0.8310 \\\scriptsize$\pm$0.0964 }  \\
\bottomrule
\end{tabular}
}
\end{sidewaystable}

\begin{sidewaystable}[]
\centering
\footnotesize
\resizebox{1.05\textheight}{!}{
\begin{tabular}{@{}cccccccccccccccccccccccc@{}}
\toprule
\multicolumn{2}{c}{Models}                                             & \multicolumn{2}{c}{FEDformer} & \multicolumn{2}{c}{Pyraformer} & \multicolumn{2}{c}{Autoformer} & \multicolumn{2}{c}{N-BEATS} & \multicolumn{2}{c}{Informer} & \multicolumn{2}{c}{LogTrans} & \multicolumn{2}{c}{Reformer} & \multicolumn{2}{c}{Transformer} & \multicolumn{2}{c}{DeepAR} & \multicolumn{2}{c}{LSTNet}  & \multicolumn{2}{c}{TCN} \\ \midrule
\multicolumn{2}{c}{MSE}                                             & base           & our           & base         & our         & base           & our           & base          & our         & base          & our          & base          & our          & base          & our          & base            & our           & base         & our         & base         & our        & base        & our       \\ \midrule

% ETTh1
\multicolumn{1}{l|}{} & \multicolumn{1}{l|}{48}  & \makecell{ 0.0509 \\\scriptsize$\pm$0.0010 } & \makecell{ \textbf{0.0490} \\\scriptsize$\pm$0.0008 } & \makecell{ 0.1485 \\\scriptsize$\pm$0.0149 } & \makecell{ 0.1144 \\\scriptsize$\pm$0.0079 } & \makecell{ 0.0760 \\\scriptsize$\pm$0.0134 } & \makecell{ 0.0658 \\\scriptsize$\pm$0.0057 } & \makecell{ 0.1337 \\\scriptsize$\pm$0.0290 } & \makecell{ 0.0963 \\\scriptsize$\pm$0.0230 } & \makecell{ 0.2216 \\\scriptsize$\pm$0.1100 } & \makecell{ 0.1608 \\\scriptsize$\pm$0.0463 } & \makecell{ 0.1257 \\\scriptsize$\pm$0.0474 } & \makecell{ 0.1247 \\\scriptsize$\pm$0.0249 } & \makecell{ 0.3218 \\\scriptsize$\pm$0.1346 } & \makecell{ 0.2940 \\\scriptsize$\pm$0.1004 } & \makecell{ 0.1925 \\\scriptsize$\pm$0.1114 } & \makecell{ 0.1479 \\\scriptsize$\pm$0.0983 } & \makecell{ 0.3035 \\\scriptsize$\pm$0.0400 } & \makecell{ 0.2499 \\\scriptsize$\pm$0.0196 } & \makecell{ 0.1294 \\\scriptsize$\pm$0.0051 } & \makecell{ 0.1130 \\\scriptsize$\pm$0.0030 } & \makecell{ 0.1770 \\\scriptsize$\pm$0.1064 } & \makecell{ 0.1230 \\\scriptsize$\pm$0.0264 }   \\
\bottomrule
\end{tabular}
}
% }
\end{sidewaystable}




% LNCS Homepage, \url{http://www.springer.com/lncs}. Last accessed 4
% \end{thebibliography}
\end{document}